\documentclass[review,3p,doublespace,authoryear]{elsarticle}

\usepackage{amssymb}

\usepackage{amsfonts}
\usepackage{amsmath}
\usepackage{amsthm}
\usepackage{color}
\usepackage[T1]{fontenc}
\usepackage{graphicx}
\usepackage{pgfplots}
\pgfplotsset{compat = newest}
\usepackage{hyperref}
\usepackage{mathrsfs}
\usepackage{upgreek}
\usepackage{import}
\usepackage{bm}
\usepackage{tikz}
\usepackage{tabulary}
\usepackage{tabularx}
\usepackage{colortbl}
\usepackage{placeins}
\usepackage{floatrow}
\usepackage{xfrac}

\usepackage{algorithmic}

\newcommand{\comment}[1]{\(\triangleright\) #1}

\pgfplotsset{compat=1.15}
\usetikzlibrary{shapes.arrows}
\usetikzlibrary{shapes.multipart}
\usetikzlibrary{shapes.geometric}
\usetikzlibrary{calc}
\usetikzlibrary{positioning,arrows}
\usepgfplotslibrary[groupplots]

\definecolor{clr1}{RGB}{0,84,159}       
\definecolor{clr2}{RGB}{161,16,53}      
\definecolor{clr3}{RGB}{0,97,101}       
\definecolor{clr4}{RGB}{246,168,0}      
\definecolor{clr5}{RGB}{87,171,39}      
\definecolor{clr6}{RGB}{156,158,159}    
\definecolor{clr7}{RGB}{100,101,103}    
\definecolor{clr8}{RGB}{122,111,172}    
\definecolor{clr9}{RGB}{0,152,161}      
\definecolor{cam-gray}{RGB}{133, 134, 139}  
\definecolor{cam-light-gray}{RGB}{219, 211, 216}  
\definecolor{cam-green}{RGB}{137, 186, 23}  
\definecolor{cam-blue}{RGB}{93, 169, 233}  
\definecolor{cam-red}{RGB}{232, 98, 82}  
\definecolor{cam-purple}{RGB}{127, 106, 147}  
\definecolor{cam-purple-2}{RGB}{180, 157, 197}  
\definecolor{cam-purple-3}{RGB}{121, 92, 143}  
\definecolor{cam-orange}{RGB}{243, 167, 18}  
\definecolor{cam-turquoise}{RGB}{192, 230, 222}  
\definecolor{cam-green-2}{RGB}{170, 206, 143}

\newcommand{\ce}{\bar{\bm{C}}_e}

\newcommand{\ui}{\bm{U}_p^{-1}}
\newcommand{\GAM}{\bar{\bm{\Gamma}}}
\newcommand{\bpe}{\bar{\bm{B}}_{p_e}}

\begin{document}

\begin{frontmatter}


\cortext[cor1]{Corresponding author}

\title{Accounting for plasticity: An extension of inelastic Constitutive Artificial Neural Networks}



\author[add1]{Birte Boes\corref{cor1}}
\ead{bboes@uni-wuppertal.de}
\author[add1]{Jaan-Willem Simon}
\ead{jsimon@uni-wuppertal.de}
\author[add2]{Hagen Holthusen}
\ead{holthusen@ifam.rwth-aachen.de}

\address[add1]{Computational Applied Mechanics, University of Wuppertal, Pauluskirchstrasse 7, 42285 Wuppertal, Germany}
\address[add2]{Institute of Applied Mechanics, RWTH Aachen University, Mies-van-der-Rohe-Str. 1, 52074 Aachen, Germany}

\begin{abstract}
        In this work, we extend the existing framework of inelastic constitutive artificial neural networks (iCANNs) by incorporating plasticity to increase their applicability to model more complex material behavior. The proposed approach ensures objectivity, material symmetry, and thermodynamic consistency, providing a robust basis for automatic model discovery of constitutive equations at finite strains. These are predicted by discovering formulations for the Helmholtz free energy and plastic potentials for the yield function and evolution equations in terms of feed-forward networks. Our framework captures both linear and nonlinear kinematic hardening behavior. 
        Investigation of our model's prediction showed that the extended iCANNs successfully predict both linear and nonlinear kinematic hardening behavior based on experimental and artificially generated datasets, showcasing promising capabilities of this framework. Nonetheless, challenges remain in discovering more complex yield criteria with tension-compression asymmetry and addressing deviations in experimental data at larger strains. Despite these limitations, the proposed framework provides a promising basis for incorporating plasticity into iCANNs, offering a platform for advancing in the field of automated model discovery.
\end{abstract}




\end{frontmatter}


\newpage

\section*{Notations}
\phantomsection
\label{sec:notations}

\begin{table}[t!bhp]
    \begin{tabular}{|l p{13cm} |}
    \hline
    $a$ & Scalar quantity   \\
    $\bm{a}$ & First order tensor \\
    $\bm{A}$  & Second order tensor  \\
    $\dot{\bm{A}}$ & Total derivative of $\bm{A}$ with respect to time \\
    $\bm{A}^T$ & Transpose of $\bm{A}$ \\
    $\bm{A}^{-1}$ & Inverse of $\bm{A}$ \\
    $\text{tr}(\bm{A})$ & Trace of $\bm{A}$ \\
    $\text{det}(\bm{A})$ & Determinant of $\bm{A}$  \\
    $\text{ln}(\bm{A})$ & Logarithm of $\bm{A}$ \\
    $\text{sym}(\bm{A})$ & Symmetric term of $\bm{A}$  \\
    $\bm{A} : \bm{B}$ & Scalar product of two tensors $\bm{A}$ and $\bm{B}$ \\
    $I_1^{\bm{A}}$ & $\text{tr}(\bm{A})$ \\
    $I_2^{\bm{A}}$ & $\frac{1}{2} \left( \text{tr}(\bm{A})^2 - \text{tr}(\bm{A}^2)\right)$ \\
    $I_3^{\bm{A}}$ & $\text{det}(\bm{A})$ \\
    $J_2^{\bm{A}}$ & $\frac{1}{2} \left( \text{tr}(\text{dev}(\bm{A}))^2 \right)$ \\
    $J_3^{\bm{A}}$ & $\text{det}(\text{dev}(\bm{A}))$ \\
    \hline
\end{tabular}
\end{table}
\section{Introduction}

The integration of machine learning into material modeling has rapidly advanced throughout the last years. By combining constitutive modeling with automated model discovery processes, it is aimed to simplify the time-consuming process of developing accurate material models. A key challenge in this field is predicting the stress-strain relation, which has led to growing interest in the application of neural networks to this problem. As reviewed by~\cite{linden2023neural,fuhg2024review,watson2024machine}, a variety of data-driven approaches have been proposed to capture constitutive laws describing the material behavior.\\ 
\textbf{Use of constitutive artificial neural networks.}\quad
Among these approaches, the framework of constitutive artificial neural networks (CANNs), initially presented by~\cite{linka2021constitutive} and~\cite{linka2023new}, has gained significant attention. This framework is distinguished by its architecture, which inherently satisfies the principles of continuum mechanics and thermodynamic consistency. As it is by no means the only model incorporating physical constraints~(see for example~\cite{linden2023neural,flaschel2023automated,masi2021thermodynamics} among many others), it has found wide acceptance and has been successfully applied in diverse fields such as Alzheimer's disease research~\citep{zhang2024discovering,stockman2024two}, human tissues, such as brain cortices~\citep{linka2023bautomated,hou2024automated}, cardiac tissues~\citep{martonova2024automated,vervenne2024constitutive}, and human skin~\citep{linka2023automated}. Additionally, applications extend to the mechanics of artificial meat~\citep{pierre2023discovering,st2024mechanical} and textile structures~\citep{mcculloch2024automated}. In addition, it recently served as foundation for the extension to Kolmogorov Arnold networks. The variety of these applications highlight its versatility and strength.\\  
While automated model discovery has been shown to be a powerful tool in elasticity, extending these approaches to inelasticity is crucial for capturing realistic material behavior. 
\cite{abdolazizi2023viscoelastic} recently extended CANNs to visco-elasticity using a Prony series, while~\cite{holthusen2023theory} introduced a general formulation for inelasticity through iCANNs. This extension broadens the framework's applicability to a wide range of material behaviors, including for example visco-elasticity, growth modeling~\citep{holthusen2025automated,holthusen2024polyconvex}, or plasticity.\\
\textbf{Plasticity in neural networks.}\quad
Since plasticity significantly influences the durability and consequently the applicability of materials extending the framework in the context of predicting yield criteria and evolution laws was of high interest recently. Approaches that employ neural networks for predicting yield criteria have been explored by~\cite{settgast2020hybrid,malik2021hybrid,vlassis2021component,vlassis2021sobolev}, while enhancing phenomenological yield functions with machine learning has been demonstrated by~\cite{fuhg2023enhancing}. Further advancements in predicting complex material responses and evolution laws are documented by~\cite{fuhg2023modular,meyer2023thermodynamically,nascimento2023machine}. Moreover, elasto-viscoplasticity has been incorporated into physics-informed neural networks, as explored by~\cite{eghbalian2023physics,eghtesad2024nn} and~\cite{keshavarz2025advancing}, while constitutive models for plasticity have been further developed by~\cite{tancogne2021recurrent,ibragimova2022convolutional,heidenreich2024recurrent,heidenreich2024transfer,shang2024analysis,weber2023physically}. 
Additionally, plasticity within artificial neural networks has been explored by~\cite{ali2019application,huang2020machine} and more recently by~\cite{ebrahim2024artificial,wang2025gpm}. Efforts to improve predictions of hardening behavior~\citep{zhang2020using,li2022counterexample,flaschel2022discovering} and advancements in non-associative plasticity~\citep{xu2025discovering} further demonstrate the field's progress. Recently, work on benchmarks to evaluate neural networks capturing elasto-plastic behavior has been described~\citep{lesueur2025benchmark}. Further, neural networks have also been applied to describe microstructure interactions~\citep{weng2023quantitative,heidenreich2023modeling,hu2024temporal} and to crystal plasticity, as discussed by~\cite{Bonatti2022,bonatti2022importance,de2022predicting,zhou2024physics}.\\ 
Given the rapid growth of automated model discovery and the increasing focus on plasticity, incorporating plasticity into iCANNs offers a promising path towards a wide application of this framework. Building on this, we aim to incorporate plasticity into the framework of iCANNs that inherently fulfills thermodynamic consistency such that physically reasonable material models are found. In contrast to the general discussion on inelasticity, we include prediction of the yield surface, hardening effects, and a return-mapping algorithm. Recently,~\cite{jadoon2025automated} described an approach similar to the one presented herein which underlines the significance of the work. However, our approach is capable to predict the overall material model including not only the discovery of the plastic components but also the overall stress response such that the total material behavior is automatically discovered.\\
\textbf{Outline}.\quad
We provide the theoretical derivation of the constitutive modeling of plasticity including linear and nonlinear kinematic hardening in Section~\ref{sec:2}. It is derived in a co-rotated intermediate configuration simplifying the implementation. The network formulation and architecture is discussed in Section~\ref{sec:3}, giving insights into the requirements of the architecture. We propose a formulation of the Helmholtz free energies to account for elastic material behavior and hardening effects as well as plastic potentials describing the yield surface and evolution equations. The automated model discovery is described in Section~\ref{sec:5}, giving details on the implementation, specific architecture and prediction performance, using artificially generated and experimental data.
\section{Constitutive Modeling} \label{sec:2}

In the following, the constitutive framework is presented, which is the basis for our network architecture. The characteristic equations are derived following the principles of thermodynamics such that a thermodynamically consistent formulation is found. 
\subsection{Kinematics}
We consider finite strains and apply the multiplicative decomposition of the deformation gradient into an elastic and plastic component according to $\bm{F}=\bm{F}_e \bm{F}_p$~\citep{lee1967finite,lee1969elastic}. To incorporate nonlinear kinematic hardening into the model formulation, we perform an additional multiplicative split of the plastic deformation gradient, i.e. $\bm{F}_p = \bm{F}_{p_e} \bm{F}_{p_i}$, which introduces a physically motivated additional intermediate configuration~\citep{lion2000constitutive}.\\
To derive a thermodynamically consistent model formulation for inelastic materials, we formulate a Helmholtz free energy that must fulfill the requirements of objectivity, material symmetry, and rotational non-uniqueness. Therefore, we define it in terms of the elastic right Cauchy-Green tensor, $\bm{C}_e=\bm{F}_e^T\bm{F}_e$, and the inelastic left Cauchy-Green tensors, $\bm{B}_{p}=\bm{F}_{p}\bm{F}_{p}^T$ and $\bm{B}_{p_e}:=\bm{F}_{p_e}\bm{F}_{p_e}^T$, i.e. $\psi = \hat{\psi} ( \bm{C}_e, \bm{B}_p, \bm{B}_{p_e})$


\textbf{Co-rotated intermediate configuration}\quad
The multiplicative split is associated with a fictitious, unstressed intermediate configuration that suffers from a rotational non-uniqueness. Thus, neither stress- nor strain-like quantities in the intermediate configuration can be computed as they suffer from the rotational non-uniqueness and in general, pull-back operations are required. Following the concept presented by~\cite{holthusen2022inelastic} and applied by~\cite{boes2023novel}, a co-rotated intermediate configuration is introduced as a co-rotated pull-back from the intermediate configuration. The eigenvalues and symmetry properties between the intermediate configuration and the co-rotated one remain the same, and the assumption of an unstressed intermediate configuration holds true here. Consequently, we can introduce the co-rotated Cauchy-Green tensors
\begin{align}
    &\ce := \bm{R}_p^{-1} \bm{C}_e \bm{R}_p = \ui \bm{C} \ui \label{Eq-BarCe}\\
    &\bar{\bm{B}}_p := \bm{R}_p^{-1} \bm{B}_p \bm{R}_p \equiv \bm{C}_p \label{Eq-BarBp} \\
    &\bar{\bm{B}}_{p_e} := \bm{R}_p^{-1} \bm{B}_{p_e} \bm{R}_p = \bm{U}_p \bm{C}^{-1}_{p_i} \bm{U}_p = \bm{U}_p \bm{U}_{p_i}^{-2} \bm{U}_p \label{Eq-BarBpe}.
\end{align} 
Following these arguments, we can describe the Helmholtz free energy by $\psi = \hat{\psi}(\ce,\bm{C}_p, \bar{\bm{B}}_{p_e})$.

\subsection{Derivation based on the Clausius-Planck inequality}
The constitutive equations of our model are derived by evaluating the Clausius-Planck inequality that reads
$-\Dot{\psi}+\bm{S} : \dfrac{1}{2}\, \Dot{\bm{C}} \ge 0$.
Therefore, we insert the time derivative of the Helmholtz free energy  
\begin{align}
    \dot{\psi} = \dfrac{\partial \psi}{\partial \ce} : \dot{\bar{\bm{C}}}_e + \dfrac{\partial \psi}{\partial \bm{C}_p} : \dot{\bm{C}}_p + \dfrac{\partial \psi}{\partial \bar{\bm{B}}_{p_e}} : \dot{\bar{\bm{B}}}_{p_e},
\end{align}
with the rates
\begin{align}
&\Dot{\bar{\bm{C}}}_e =\ui \Dot{\bm{C}} \ui - \ce \bar{\bm{L}}_p - \bar{\bm{L}}_p^T \ce \\
&\Dot{\bm{C}}_p = \bar{\bm{L}}_p^T \bm{C}_p + \bm{C}_p \bar{\bm{L}}_p \\
&\Dot{\bm{B}}_{p_e} = - \bm{U}_p \bm{C}_{p_i}^{-1} \Dot{\bm{C}}_{p_i} \bm{C}_{p_i}^{-1}  \bm{U}_p + \bar{\bm{L}}_p \bar{\bm{B}}_{p_e} + \bar{\bm{B}}_{p_e} \bar{\bm{L}}^{T}_p
\end{align}
where $\bar{\bm{L}}_p=\Dot{\bm{U}}_p\ui$ and $\bar{\bm{L}}_{p_i}=\Dot{\bm{U}}_{p_i}\bm{U}_{p_i}$ holds. Using $\bar{\bm{D}}_p := \text{sym}(\bar{\bm{L}}_p)$ and $\bar{\bm{D}}_{p_i} := \text{sym}(\bar{\bm{L}}_{p_i})$, the dissipation inequality reads 
\begin{align}
    \left( \bm{S} - 2\,\bm{U}_p^{-1} \dfrac{\partial \psi}{\partial \bar{\bm{C}}_e} \bm{U}_p^{-1}\right) : \dfrac{1}{2}\dot{\bm{C}} 
    + \underbrace{\left( \underbrace{ 2\,\bar{\bm{C}}_e \dfrac{\partial \psi}{\partial \bar{\bm{C}}_e}}_{=:\bar{\bm{\Sigma}}} 
    - \underbrace{2 \,\dfrac{\partial \psi}{\partial {\bm{C}}_p} {\bm{C}}_p}_{=:\bar{\bm{\chi}}}
    - \underbrace{2 \,\dfrac{\partial \psi}{\partial \bar{\bm{B}}_{p_e}} \bar{\bm{B}}_{p_e}}_{=:\bar{\bm{\Xi}}}\right) }_{=:\bar{\bm{\Gamma}}}
     : \bar{\bm{D}}_p \nonumber \\
    + \, \underbrace{ 2 \,\bm{U}_{p_i}^{-1} \bm{U}_p \dfrac{\partial \psi}{\partial \bar{\bm{B}}_{p_e}} \bm{U}_p \bm{U}_{p_i}^{-1}}_{=: \bar{\bm{\Theta}}} : \bar{\bm{D}}_{p_i}
     \ge 0. \label{Eq-CDI}
\end{align}
Applying the procedure by~\cite{coleman1961foundations}, we get the second Piola-Kirchhoff stress, $\bm{S}$,
and define the state relations of the relative stress, $\bar{\bm{\Gamma}}:=\bar{\bm{\Sigma}}-\bar{\bm{\chi}}- \bar{\bm{\Xi}}$, defined by the Mandel stress, $\bar{\bm{\Sigma}}$, the backstresses, $\bar{\bm{\chi}}$ and $\bar{\bm{\Xi}}$, and the Mandel stress corresponding to kinematic hardening, $\bar{\bm{\Theta}}$. The backstresses are related to kinematic hardening and describe the translation of the origin of the yield surface during plastic deformations. 
All quantities are listed in Table~\ref{tab:Const-rel} for an overview. 
While neither the backstresses nor the Mandel stress are generally symmetric, the relative stress is (cf.~\cite{svendsen2001modelling}). Consequently, we have used the relations $\bar{\bm{D}}_p := \text{sym}(\Dot{\bm{U}}_p \bm{U}_p^{-1})$ and $\bar{\bm{D}}_{p_i} := \text{sym}(\Dot{\bm{U}}_{p_i} \bm{U}_{p_i}^{-1})$.
From this, we obtain the reduced dissipation inequality, i.e.
\begin{align}
    \mathcal{D}_\text{red} := \bar{\bm{\Gamma}} : \bar{\bm{D}}_p + \bar{\bm{\Theta}} : \bar{\bm{D}}_{p_i} \ge 0. \label{Eq-ReducedInequality}
\end{align}
For more detailed derivations, the interested reader is referred to~\cite{vladimirov2008modelling,brepols2020gradient,felder2020incorporating,holthusen2022inelastic}.
The reduced dissipation inequality must be fulfilled for arbitrary thermodynamical processes. Therefore, we introduce evolution equations for the strain-like quantities in terms of plastic potentials, which we describe by scalar-valued isotropic functions of the thermodynamic driving forces, $g_1 = \hat{g_1}(\bar{\bm{\Gamma}})$ and $g_2 = \hat{g_2}(\bm{\Theta})$, such that
\begin{align}
    \bar{\bm{D}}_p = \lambda \dfrac{\partial g_1}{\partial \bar{\bm{\Gamma}}} \quad &\rightarrow \quad \Dot{\bm{C}}_p = 2 \lambda \bm{U}_p \dfrac{\partial g_1}{\partial \bar{\bm{\Gamma}}} \bm{U}_p \label{Eq-Dp} \\
    \bar{\bm{D}}_{p_i} = \lambda \dfrac{\partial g_2}{\partial \bar{\bm{\Gamma}}} \quad &\rightarrow \quad \Dot{\bm{C}}_p = 2 \lambda \bm{U}_{p_i} \dfrac{\partial g_2}{\partial \bar{\bm{\Theta}}} \bm{U}_{p_i}. \label{Eq-Dpi}
\end{align}   
The latter one can be used to account for nonlinear kinematic hardening of Frederick-Armstrong type~\citep{armstrong1966mathematical}. Further, we restrict the potentials to be convex, zero-valued and non-negative, automatically fulfilling Inequality~(\ref{Eq-ReducedInequality})~(see~\cite{holthusen2025automated}).
We interpret $\lambda$ as a plastic multiplier, considering inelastic effects of plasticity. Regarding a wider class of inelasticity, it can be interpreted, for example, as a relaxation time in visco-elasticity or a growth multiplier for growth modeling (c.f.~\cite{lamm2021macroscopic,lamm2022macroscopic}). The plastic
multiplier is (per definition) non-negative and is defined through the Karush-Kuhn-Tucker conditions, that read
\begin{align}
    \lambda \ge 0, \quad \Phi \le 0, \quad\lambda \,\Phi = 0. \label{Eq-KKT}
\end{align}
They complete the set of constitutive equations and introduce the yield function $\Phi$, which encloses the set of all admissible stress states. All constitutive equations are summarized in Table~\ref{tab:Const-rel}. Their specific formulation in terms of the Helmholtz free energy, $\psi$, and plastic potentials, $g$, will be described in detail in Sections~\ref{sec:NN-psi} and~\ref{sec:NN-g}.

\begin{table}[bth]
    \begin{tabular}{l l}
        \hline
        Constitutive equations \\ \hline 
        \textbf{Stresses} & \\
        Second Piola-Kirchhoff stress & $\bm{S} = 2\,\bm{U}_p^{-1} \dfrac{\partial \psi}{\partial \bar{\bm{C}}_e} \bm{U}_p^{-1}$ \\[0.2cm]
        Mandel stress & $\bar{\bm{\Sigma}} =  2\,\bar{\bm{C}}_e \dfrac{\partial \psi}{\partial \bar{\bm{C}}_e}$ \\[0.2cm]
        Linear backstress & $\bar{\bm{\chi}}= 2 \,\dfrac{\partial \psi}{\partial {\bm{C}}_p} {\bm{C}}_p$ \\[0.25cm]
        Nonlinear backstress & $\bar{\bm{\Xi}} =  2 \,\dfrac{\partial \psi}{\partial \bar{\bm{B}}_{p_e}} \bar{\bm{B}}_{p_e}$ \\[0.2cm]
        Relative stress& $\bar{\bm{\Gamma}}=\bar{\bm{\Sigma}}-\bar{\bm{\chi}}-\bar{\bm{\Xi}}$ \\[0.2cm]
        Mandel stress related to kinematic hardening & $\bar{\bm{\Theta}} =  2 \,\bm{U}_{p_i}^{-1} \bm{U}_p \dfrac{\partial \psi}{\partial \bar{\bm{B}}_{p_e}} \bm{U}_p \bm{U}_{p_i}^{-1}$\\
        \textbf{Evolution Equations} & 
         $\Dot{\bm{C}}_p = 2 \, \lambda\, \bm{U}_p \dfrac{\partial g_1}{\partial \bar{\bm{\Gamma}}} \bm{U}_p$ \\[0.2cm]
         &$\Dot{\bm{C}}_{p_i} = 2\, \lambda \,\bm{U}_{p_i} \dfrac{\partial g_2}{\partial \bar{\bm{\Gamma}}} \bm{U}_{p_i}$ \\[0.2cm]
        \textbf{Yield criterion} &
        $\Phi(\bm{\bar{\Gamma}}) = g_{\Phi}(\bar{\bm{\Gamma}}) - 1$ \\ \hline
    \end{tabular}
    \caption{Summary of all constitutive relations.}\label{tab:Const-rel}
\end{table}
\section{Network Formulation and Architecture}\label{sec:3}

So far, the constitutive equations have been defined in a general manner. It remains to find specific formulations for the Helmholtz free energy, $\psi$, the yield function, $\Phi$, and the potentials, $g_1$ and $g_2$. 
While a specific formulation is found in classical constitutive modeling, we define them in terms of feed-forward networks such that the final form can be automatically discovered.
The overall framework is built in analogy to the work presented in~\cite{holthusen2023theory}. In the following, we outline the key aspects of our architecture and highlight the novel extensions related to plasticity. 

\subsection{Architecture}
\begin{figure}
    \includegraphics{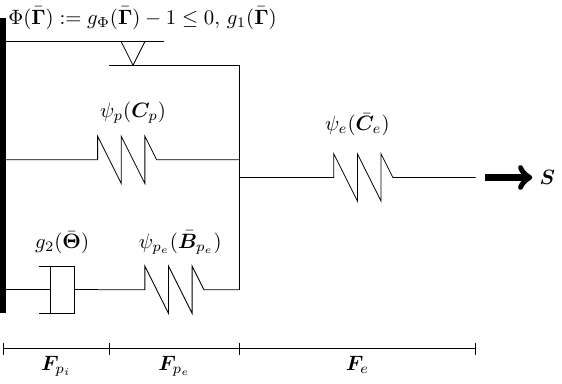}
    \caption{Rheological representation of the iCANN capturing plasticity consisting out of three springs, $\psi_e$, $\psi_p$, and $\psi_{p_e}$. A damper, $\Phi$ and $g_1$, and a dashpot element, $g_2$, with the states $\bm{U}_p$ and $\bm{U}_{p_i}$, are connected in parallel.}~\label{Fig:rheolog-model}
\end{figure}

All feed-forward networks are embedded in a recurrent neural network which is evaluated at each time step.  The structure is visualized as a flowchart in Figure~\ref{Fig:RNN} and will be described in detail in the following. Strain quantities serve as inputs, while the computed stresses serve as outputs. The latter ones are used as loss measure where we compute the mean squared error between the experimentally observed and predicted stresses. 
All history variables are passed through the hidden states. In total, the architecture comprises the following components:
\begin{itemize}
    \item $\psi_e$~--~elastic Helmholtz free energy
    \item $\psi_p$~--~Helmholtz free energy to account for linear kinematic hardening 
    \item $\psi_{p_e}$~--~Helmholtz free energy to account for nonlinear kinematic hardening 
    \item $g_{\Phi}$~--~plastic potential defining the stress-dependent part of the yield surface 
    \item $g_1$~--~plastic potential defining the evolution of $\bm{C}_p$
    \item $g_2$~--~plastic potential defining the evolution of $\bpe$
\end{itemize}
A rheological representation is given in Figure~\ref{Fig:rheolog-model}, where a parallel connection of a dashpot, a spring, and a damper in series with a spring represent the plastic material answer in combination with linear and nonlinear kinematic hardening.
The total Helmholtz free energy is defined by $\psi := \psi_e + \psi_p + \psi_{p_e}$.
For associative plasticity, we obtain $g_{\Phi}=g_1$, reducing the number of FFNs by one. Excluding hardening effects further reduces the amount of trainable networks.
\\
We use explicit time integration for the evolution equations in Equations~(\ref{Eq-Dp}) and~(\ref{Eq-Dpi}). As visualized in Figure~\ref{Fig:RNN}, first, we compute all kinematic, stress and stress-like quantities from the last time step $t_n$ (upper gray box) by evaluating the Helmholtz free energies. These are described by FFNs which comprise the same architecture, visualized by orange boxes. Next, we evaluate the plastic potentials, $g_{1_n}:=g_1(t=t_n)$ and $g_{2_n}:=g_2(t=t_n)$, to update the explicit evolution equations. These are again formulated by FFNs and colored by blue boxes to distinguish between the different architectures of the FFNs for the Helmholtz free energy and the potentials. We then compute all kinematic quantities from the current time step (lower gray box), $t_{n+1}$, and determine the current stress and stress-like quantities by evaluating the Helmholtz free energy (orange boxes). As we consider plasticity, we enforce the Karush-Kuhn-Tucker conditions in Equation~(\ref{Eq-KKT}) to be fulfilled. Thus, we employ an elastic trial step and, if necessary, update the plastic multiplier iteratively until convergence is achieved. Therefore, the FFN of the yield function (blue box) is called. Finally, we give the second Piola-Kirchhoff stress as an output and compute the corresponding Cauchy stress by 
\begin{align}
    \bm{\sigma} = \text{det}(\bm{F})^{-1} \bm{F} \bm{S} \bm{F}^T. \label{Eq-Cauchy}
\end{align}
    \begin{figure}[h!t]
        \centering
        \vspace*{-1.5cm}
        \begin{tikzpicture}[scale=0.85]
            
            \tikzstyle{io} = [trapezium, trapezium left angle=70, trapezium right angle=110, minimum width=1.25cm, minimum height=0.25cm, text centered, draw=black, fill=clr1!15]
            \tikzstyle{process} = [rectangle, minimum width=1.5cm, minimum height=0.55cm, text centered, draw=black, fill=clr6!5]
            \tikzstyle{FFN} = [rectangle, minimum width=1.75cm, minimum height=1cm, text centered, draw=clr4,rounded corners=0.1cm, fill=clr4!15]
            \tikzstyle{FFN-g} = [rectangle, minimum width=1.75cm, minimum height=1cm, text centered, draw=clr9,rounded corners=0.1cm, fill=clr9!15]
            \tikzstyle{decision} = [diamond, minimum width=2cm, minimum height=1cm, text centered, draw=black, fill=clr6!30]
            \tikzstyle{arrow} = [thick,->,>=stealth]

            \draw[draw=clr6,fill=clr6!3,line width=0.5mm, rounded corners=0.1 cm] (-5.8,10.35) rectangle (6.5,-10.45);
            \draw[draw=clr7,fill=clr6!15, rounded corners=0.1 cm] (-5.6,10.2) rectangle (6.3,1.42);      
            \draw[draw=clr7,fill=clr6!15, rounded corners=0.1 cm] (-5.6,1.125) rectangle (6.3,-10.3); 

            \node[clr7] at (-5.2,9.85) {$t_n$};
            \node[clr7] at (-5.0,0.75) {$t_{n+1}$};

            \node[] at (0,10.8) [io] (IN)  { \small $\bm{C}_n,\, \bm{U}_{p_n}, \bm{U}_{{p_i}_n}$};
            \node[left=0.2cm of IN.west, anchor=east] (states) { \small States:};

            \node[below=0.65cm of IN.south] (A2) [process]  { \small ${\bm{C}_p}_n$};
            \node[left=1.5cm of A2.west] (A1) [process]  { \small $\bar{\bm{C}}_{e_n}$};
            \node[right=1.5cm of A2.east] (A3) [process]  { \small $\bar{\bm{B}}_{{p_e}_n}$};
            \node[clr7,right=1.7cm of A3.east, anchor = west] (Eq1) {\small \comment{Eq.~(\ref{Eq-BarCe}) -~(\ref{Eq-BarBpe})}};

            \node[below=0.5cm of A1.south] (B1) [FFN]  { \small $\text{FFN}_{\psi_e}$};
            \node[below=0.5cm of A2.south] (B2) [FFN]  { \small $\text{FFN}_{\psi_p}$};
            \node[below=0.5cm of A3.south] (B3) [FFN]  { \small $\text{FFN}_{\psi_{p_e}}$};
            \node[clr7,right=1.6cm of B3.east, anchor = west] (Eq1) {\small \comment{Eq.~(\ref{Eq-Psi}), Fig.~\ref{Fig-Psi-e}}};

            \node[below=0.5cm of B1.south] (C1) [process]  { \small $\bar{\bm{\Sigma}}_n$};
            \node[below=0.5cm of B2.south] (C2) [process]  { \small $\bar{\bm{\chi}}_n$};
            \node[below=0.5cm of B3.south] (C3) [process]  { \small $\bar{\bm{\Xi}}_n,\,\bm{\Theta}_n$};            
            \node[clr7,right=1.7cm of C3.east, anchor = west] (Eq1) {\small \comment{Table~\ref{tab:Const-rel}}};

            \node[below=0.5cm of C2.south] (D1) [process]  { \small $\bar{\bm{\Gamma}}_n$};
            \node[clr7,right=4.7cm of D1.east, anchor = west] (Eq1) {\small \comment{Table~\ref{tab:Const-rel}}};

            \node[below=0.5cm of D1.south] (E2) [FFN-g]  { \small $\text{FFN}_{g_1}$};
            \node[right=1.25cm of E2.east] (E1) [FFN-g]  { \small $\text{FFN}_{g_2}$};
            \node[clr7,right=1.6cm of E1.east, anchor = west] (Eq1) {\small \comment{Eq.~(\ref{Eq-g-initial}), Fig.~\ref{Fig-potential}}};
            
            \node[below=0.5cm of E1.south] (F1) [process]  { \small $\bm{D}_{p_i}$};
            \node[below=0.5cm of E2.south] (F2) [process]  { \small $\bm{D}_{p}$};
            \node[clr7,right=1.7cm of F1.east, anchor = west] (EqD) {\small \comment{Eq.~(\ref{Eq-Dp}) -~(\ref{Eq-Dpi})}};
            
            \node[left=5cm of F2.south west, anchor=north east,yshift=-0.59mm] (Cn1) [io]  { \small $\bm{C}$};
            \node[left=0.2cm of Cn1.west, anchor=east] (Inputs) { \small Input:};

            \node[below=0.8cm of F2.south] (lambda) [process]  { \small $\lambda=0$};

            \node[below=0.5cm of lambda.south] (G2) [process]  { \small $\bm{C}_p$};
            \node[left=1.5cm of G2.west] (G1) [process]  { \small $\ce$};
            \node[right=1.5cm of G2.east] (G3) [process]  { \small $\bpe$};
            \node[clr7,right=1.7cm of G3.east, anchor = west] (Eq1) {\small \comment{Eq.~(\ref{Eq-BarCe}) -~(\ref{Eq-BarBpe})}};

            \node[below=0.5cm of G1.south] (H1) [FFN]  { \small $\text{FFN}_{\psi_e}$};
            \node[below=0.5cm of G2.south] (H2) [FFN]  { \small $\text{FFN}_{\psi_p}$};
            \node[below=0.5cm of G3.south] (H3) [FFN]  { \small $\text{FFN}_{\psi_{p_e}}$};
            \node[clr7,right=1.6cm of H3.east, anchor = west] (Eq1) {\small \comment{Eq.~(\ref{Eq-Psi}), Fig.~\ref{Fig-Psi-e}}};

            \node[below=0.5cm of H1.south] (I1) [process]  { \small $\bm{S},\,\bar{\bm{\Sigma}}$};
            \node[below=0.5cm of H2.south] (I2) [process]  { \small $\bar{\bm{\chi}}$};
            \node[below=0.49cm of H3.south] (I3) [process]  { \small $\bar{\bm{\Xi}},\,\bm{\Theta}$};
            \node[clr7,right=1.7cm of I3.east, anchor = west] (Eq1) {\small \comment{Table~\ref{tab:Const-rel}}};

            \node[below=0.5cm of I2.south] (J1) [process]  { \small $\bar{\bm{\Gamma}}$};
            \node[clr7,right=4.7cm of J1.east, anchor = west] (Eq1) {\small \comment{Table~\ref{tab:Const-rel}}};

            \node[below=0.5cm of J1.south] (K1) [FFN-g]  { \small $\text{FFN}_{g_\Phi}$};
            \node[clr7,right=4.6cm of K1.east, anchor = west] (Eq1) {\small \comment{Eq.~(\ref{Eq-g-initial}), Fig.~\ref{Fig-potential}}};

            \node[below=0.5cm of K1.south] (Phi) [decision]  { \small $\Phi \le 0$};
            \node[clr7,right=4.475cm of Phi.east, anchor = west] (Eq1) {\small \comment{Eq.~(\ref{Eq-Yield}), Eq.~\ref{Eq-lambda}}};

            \node[below=0.9cm of Phi.south] (S) [io]  { \small $\bm{S}$};
            \node[below=0.25cm of S.south] (sigma) [io]  { \small $\bm{\sigma}$};
            \node[clr7,right=0.2cm of sigma.east, anchor = west] (Eq1) {\small \comment{Eq.~(\ref{Eq-Cauchy})}};
            \node[left=0.2cm of sigma.west,anchor = east] (Output)  { \small Output:};  

            \node[right=2cm of Phi.east] (Ln1) [process] {Update $\Delta \lambda$};

            
            \draw[arrow] (IN.south) -- ++(0,-0.5) -| (A1.north);
            \draw[arrow] (IN.south) -- ++(0,-0.5) -| (A2.north);
            \draw[arrow] (IN.south) -- ++(0,-0.5) -| (A3.north);

            \draw[arrow] (A1.south) -- (B1.north);
            \draw[arrow] (A2.south) -- (B2.north);
            \draw[arrow] (A3.south) -- (B3.north);
            \draw[arrow] (B1) -- (C1);
            \draw[arrow] (B2) -- (C2);
            \draw[arrow] (B3) -- (C3);
            \draw[arrow] (C1.south) |- ++(0,-0.2) -| (D1.north);
            \draw[arrow] (C2.south) |- ++(0,-0.2) -| (D1.north);
            \draw[arrow] ([xshift=-8pt]C3.south) |- ++(0,-0.2) -| (D1.north);
            \draw[arrow] (D1) -- (E2);
            \draw[arrow] ([xshift=8pt]C3.south) -- ([xshift=8pt]E1.north);
            \draw[arrow] (E1) -- (F1);
            \draw[arrow] (E2) -- (F2);
            \draw[arrow] (Cn1.east) -| ([xshift=-5pt]lambda.north);
            \draw[arrow] (F2) -- (lambda);
            \draw[arrow] (F1.south) |- ++(0,-0.2) -| (lambda);
            \draw[arrow] (lambda.south) -- ++(0,-0.2) -| (G1);
            \draw[arrow] (lambda.south) -- ++(0,-0.2) -| (G2);
            \draw[arrow] (lambda.south) -- ++(0,-0.2) -| (G3);
            \draw[arrow] (G1) -- (H1);
            \draw[arrow] (G2) -- (H2);
            \draw[arrow] (G3) -- (H3);
            \draw[arrow] (H1) -- (I1);
            \draw[arrow] (H2) -- (I2);
            \draw[arrow] (H3) -- (I3);
            \draw[arrow] ([xshift=8pt]I1.south) -- ++(0,-0.2) -| (J1);
            \draw[arrow] (I2.south) -- ++(0,-0.2) -| (J1);
            \draw[arrow] (I3.south) -- ++(0,-0.2) -| (J1);
            \draw[arrow] (J1) -- (K1);
            \draw[arrow] (K1) -- (Phi);
            \draw[thick] ([xshift=-8pt]I1.south) |- ([yshift=0.8cm]S.north);
            \draw[arrow] (Phi) -- (S) node[pos=0.3,right] {Yes};
            \draw[arrow] (Phi.east) --  node[pos=0.5, above] {No} (Ln1.west);    
            \draw[arrow] ([xshift=1cm]Ln1.north)  |- (lambda.east);    
            \draw[arrow] (S) -- (sigma);

        \end{tikzpicture}
        \caption{Flowchart of the recurrent neural network (outer gray box) called at each time step. First, all constitutive equations from the last time step $t_n$ (upper gray box) are computed. Internal variables are explicitly updated to compute all constitutive equations of the current time step, $t_{n+1}$ (lower gray box). Feed-forward networks of the Helmholtz free energies are indicated by orange boxes, while the ones of the plastic potentials are shown by blue boxes. Inputs, outputs and states are visualized by blue trapezoids.}~\label{Fig:RNN}
    \end{figure}

\subsubsection{Algorithmic treatment}
The evolution equations are discretized within a time interval $t \in [t_{n},t_{n+1}]$, and an explicit time integration scheme is used. For this, an exponential integration scheme is applied, such that Equations~(\ref{Eq-Dp}) and~(\ref{Eq-Dpi}) yield 
\begin{align}
    &\bm{C}_{p_{n+1}} = \bm{U}_{p_n}\, \text{exp}(2\, \Delta \lambda \,\bar{\bm{D}}_p ) \,\bm{U}_{p_n}\\
    &\bm{C}_{{p_i}_{n+1}} = \bm{U}_{{p_i}_n}\, \text{exp}(2\, \Delta \lambda \,\bar{\bm{D}}_{p_i} ) \,\bm{U}_{{p_i}_n}
\end{align}
with $\Delta \lambda := \Delta t \, \lambda$ and $\Delta t= t_{n+1} - t_n$. This formulation ensures plastic incompressibility.\\
Unlike CANNs or iCANNs used for visco-elasticity, we must fulfill the Karush-Kuhn-Tucker conditions in Equation~(\ref{Eq-KKT}) such that we apply a predictor-corrector scheme by using a Newton-Raphson iteration. Thus, we introduce a residual $r=\Phi \overset{!}{=}0$, which is minimized iteratively until it converges to zero by updating the plastic multiplier by 
\begin{align}
    \Delta \lambda = -(\frac{\partial r}{\partial \lambda})^{-1}\, r \label{Eq-lambda}
\end{align}
until convergence is achieved, ensuring that the current stress state lies within the set of admissible stresses defined by the yield surface.

\subsection{Feed-forward networks}
The Helmholtz free energy and all potentials are described by feed-forward networks. Each network takes the invariants of the corresponding strain measure for the first and stress measure for the latter one as input quantities which are passed through a two-layer network.  The weights of the first layer define the shape of the activation functions while the weights of the second layer scale their contribution. All networks are not fully-connected. 
We restrict all weights to be non-negative to obtain physically reasonable results.

\subsubsection{Helmholtz free energy}\label{sec:NN-psi}
\begin{figure}[htbp]
        \includegraphics{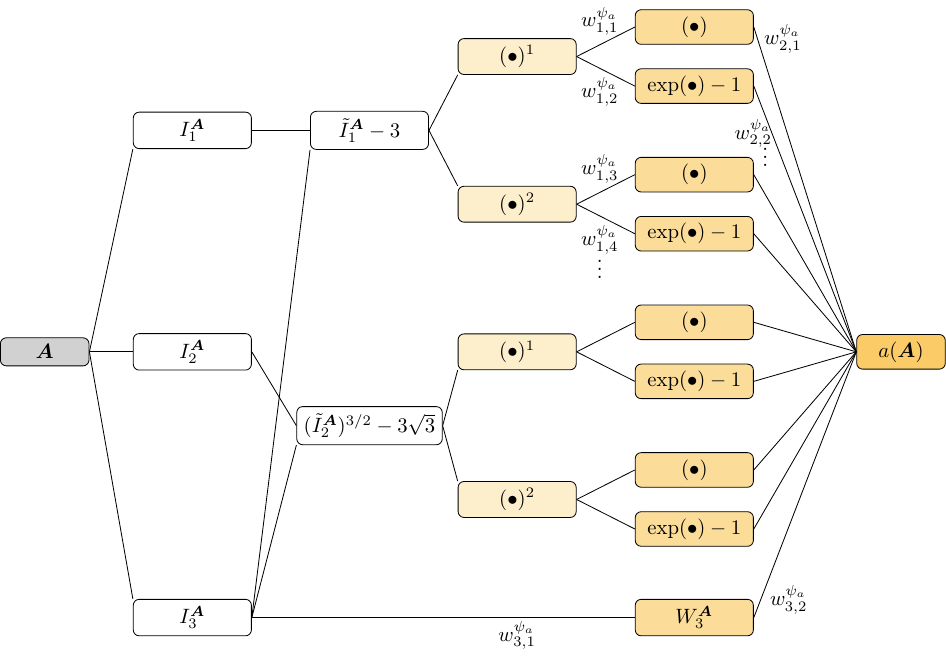}
        \caption{Schematic illustration of the feed-forward network of the elastic and plastic Helmholtz free energies. Elastic formulation: $\bm{A}=\ce$ and $a= \psi_e(\ce)$, plastic formulations: $\bm{A} = \bm{C}_p$ and $a=\psi_p(\bm{C}_p)$, as well as $\bm{A} = \bpe$ and $a=\psi_{p_e}(\bpe)$.}
        \label{Fig-Psi-e}
    \end{figure}

    \textbf{Prerequisites for the formulation.}\quad
    To define a generic formulation of the Helmholtz free energy, we define $\psi$ to be an isotropic function of its arguments and express it in terms of its invariants. Furthermore, we employ the classical isochoric-volumetric split originally introduced by~\cite{flory1961thermodynamic} to satisfy the growth criterion. In addition, we exclude all mixed invariants of $\ce$, $\bm{C}_p$, and $\bpe$ such that we can additively decompose the total Helmholtz free energy to $\psi = \psi_e(\ce) + \psi_p(\bm{C}_p) + \psi_{p_e}(\bpe)$ to easily prove polyconvexity. Resulting from this decomposition, we employ three separate networks to discover the total Helmholtz free energy. It can be interpreted as the elastic energy, $\psi_e$, and the ones accounting for linear and nonlinear kinematic hardening, $\psi_p$ and $\psi_{p_e}$ respectively.
    Together with the requirements of fulfilling the growth criterion and ensuring polyconvexity, each formulation must fulfill the normalization condition, i.e.\ being zero-valued at the origin, and ensuring a stress-free reference configuration. \\
    \textbf{Generic formulation.}\quad
    While using a feed-forward network for the elastic Helmholtz free energy has successfully been applied in CANNs and iCANNs already~\citep{linka2021constitutive,holthusen2023theory}, we extend the model to discover kinematic hardening. 
    As described above, an additive formulation of the total Helmholtz free energy is applied. Thus, we describe each subterm of $\psi$ by a separate FFN. Since each of them must fulfill the requirements given above, we can employ the same structure for all of them, which is visualized in Figure~\ref{Fig-Psi-e}. Here, $\bm{A}$ denotes the input into the network, that is either $\ce$, $\bm{C}_p$, or $\bpe$, while $a(\bm{A})$ denotes the corresponding output, that is either $\psi_e(\ce)$, $ \psi_p(\bm{C}_p)$, or $\psi_{p_e}(\bpe)$, respectively.\\
    The invariants of the related strain measure serve as input quantities and are given in the~\hyperref[sec:notations]{Notations}. As described above, we employ the isochoric-volumetric invariants: $\tilde{I}_1 = \sfrac{I_1}{I_3^{1/3}}$, $\tilde{I}_2 = \sfrac{I_2}{I_3^{2/3}}$, and $I_3$. The former ones are evaluated at the initial state, $\bm{C}=\bm{I}$, to fulfill the normalization condition and ensure convexity. Here, the second invariant has been evaluated to the power of $3/2$ to ensure convexity~\citep{hartmann2003polyconvexity,schroder2003invariant}.
    While linear and quadratic terms are applied within the first layer of the network (lighter orange layer in Figure~\ref{Fig-Psi-e}), we choose linear and exponential activation functions within the second layer of the network (darker orange layer in Figure~\ref{Fig-Psi-e}). These activation functions are monotonic, continuous, and both, continuous differentiable and zero-valued at the origin. In addition, they are unbounded such that the growth criterion is fulfilled. The volumetric term is designed following~\cite{ogden1972large}. The generic formulation of the Helmholtz free energy then reads
    \begin{align}
        \psi(\bm{A}) &= \, w_{2,1}^{\psi} \,(\tilde{I}_1^{\bm{A}}-3)+ w_{2,2}^{\psi} \left( \text{exp} (w_{1,2}^{\psi}\,[\tilde{I}_1^{\bm{A}}-3])-1\right)  \nonumber \\ \nonumber  
        &+ w_{2,3}^{\psi}\, {(\tilde{I}_1^{\bm{A}}-3)}^2      + w_{2,4}^{\psi}  \left(\text{exp} \left( w_{1,4}^{\psi}\,{[\tilde{I}_1^{\bm{A}}-3]}^2\right)-1\right) \\ \nonumber  
        &+ w_{2,5}^{\psi}\, \left((\tilde{I}_2^{\bm{A}})^{3/2}-3\sqrt{3}\right)        + w_{2,6}^{\psi}  \left(\text{exp} (w_{1,6}^{\psi}\,\left[(\tilde{I}_2^{\bm{A}})^{3/2}-3\sqrt{3}\right])-1\right)  \\ \nonumber  
        &+ w_{2,7}^{\psi}\, {\left((\tilde{I}_2^{\bm{A}})^{3/2}-3\sqrt{3}\right)}^2      + w_{2,8}^{\psi}  \left( \text{exp} \left(w_{1,8}^{\psi}\,\left[{(\tilde{I}_2^{\bm{A}})}^{3/2}-3\sqrt{3}\right]^2\right)-1\right)  \\   
        &+ w_{3,2}^{\psi} \underbrace{ \left[ {\left(I_3^{\bm{A}}\right)}^{ w_{3,1}^{\psi}} -1 -  w_{3,1}^{\psi} \text{ln} \left(I_3^{\bm{A}}\right) \right]}_{W_3^{\bm{a}}}. \label{Eq-Psi}
    \end{align}

\subsubsection{Plastic potential}\label{sec:NN-g}
    \textbf{Formulation of the yield surface in terms of a potential.}\quad
    Considering plasticity, we must formulate functions for both the yield surface and evolution equations. Focussing on the yield surface, we seek a general formulation that can be described by a network design. In general, the yield functions can be expressed in terms of a constant, which represents the initial yield stress, $\sigma_{y_0}$, and a stress-dependent term, $g_0^{\Phi}(\GAM)$, such that $\Phi_0(\GAM)= g_0^{\Phi}(\GAM) - \sigma_{y_0} \le 0$ holds. This can be reformulated to
    \begin{align}
        \Phi(\GAM) =  g_{\Phi}(\GAM) - 1, \label{Eq-Yield}
    \end{align}
    where we define $g^{\Phi} := \frac{1}{\sigma_{y_0}} g_0^{\Phi}$. Normalization with respect to the initial yield stress reduces the number of weights that need to be trained. We can deduce from this that the yield function can be described in terms of a potential $g$. Thus, we employ the same network structure that will be introduced for the potentials in Equations~(\ref{Eq-Dp}) and~(\ref{Eq-Dpi}).\\

    \textbf{Requirements on the potentials.}\quad
    The potentials are designed to be scalar-valued isotropic functions of their arguments such that we can define it in terms of stress invariants. While the choice of these invariants is not unique, we choose a formulation in terms of the integrity basis of their inputs ensuring rotational non-uniqueness. Specifically, we use the invariants $I_1$ and $J_2$ (see~\hyperref[sec:notations]{Notations}) to account for hydrostatic pressures and shear stresses. Since we additionally require the potentials to be convex, non-negative and zero-valued at the origin, we omit the third deviatoric invariant, $J_3$. Although often used in plasticity, it does not fulfill convexity in general. Further, the additionally gained information into the stress behavior is often negligible~\citep{hill1952mathematical}.\\
    We additively decompose each potential into sub-potentials that dependent separately on the invariants, i.e. $g(\bm{A}) = g_I(I_1^{\bm{A}})+g_{II}(J_2^{\bm{A}})$. Avoiding sharp corners in the intersection of the sub-functions, a smoothing technique as described in~\cite{gesto2011smoothing} such as the p-norm regularization presented by~\cite{mollica2002general} is applied. The potential then reads $  g(\GAM) = \left( \sum_{i=1}^{n} \left| g_i(\GAM) \right|^p \right)^{1/p}$ where $p$ is a positive parameter and $n$ is the number of sub-potentials. By the choice of the sub-functions $g_i$ to be convex, non-negative and zero-valued, we automatically fulfill this equation for $p=1$.

\begin{figure}[htbp]
        \includegraphics{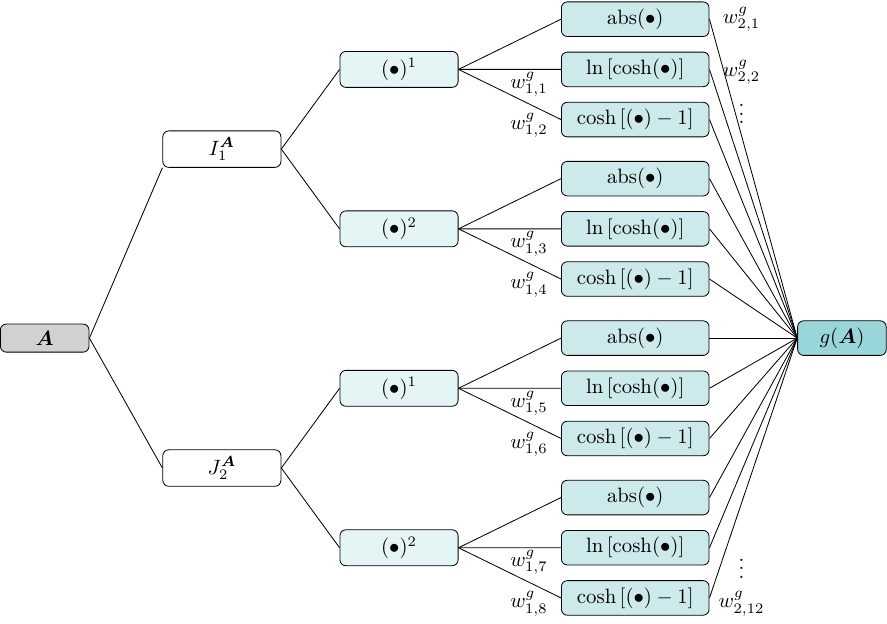}
        \caption[short]{Schematic illustration of the feed-forward network of the plastic potential of Equation~(\ref{Eq-g-initial}) with $\bm{A}=\GAM$ for $g_1$ and $g_{\Phi}$, and $\bm{A}=\bar{\bm{\Theta}}$ for $g_2$.}\label{Fig-potential}
    \end{figure}

    \textbf{Generic formulation.}\quad
    Following the requirements from above, we design a feed-forward network for the plastic potential visualized in Figure~\ref{Fig-potential}. The first layer (light blue layer) consists of linear and quadratic terms, while the activation functions in the second layer (darker blue layer) include the functions $\text{abs}(\bullet)$, $\text{ln}(\text{cosh}(\bullet))$ and $\text{cosh}(\bullet)-1$. The resulting equation for the feed-forward network therefore reads
    \begin{align}
        g(\bm{A}) &=\, w_{2,1}^{g} \,\text{abs}\left(I_1^{\bm{A}}\right) + w_{2,2}^{g} \text{ln}\left[\text{cosh}\left(w_{1,1}^{g} \,I_1^{\bm{A}}\right)\right] + w_{2,3}^{g} \left[\text{cosh}\left(w_{1,2}^{g} \,I_1^{\bm{A}}\right)-1\right] \nonumber \\
        &+ w_{2,4}^{g} \, \text{abs}\left((I_1^{\bm{A}})^2\right) + w_{2,5}^{g} \text{ln}\left[\text{cosh}\left(w_{1,2}^{g} \,(I_1^{\bm{A}})^2\right)\right]               + w_{2,6}^{g} \left[\text{cosh}\left(w_{1,4}^{g} \,(I_1^{\bm{A}})^2\right)-1\right]  \nonumber  \\
        &+ w_{2,7}^{g} \, \text{abs}\left(J_2^{\bm{A}}\right) + w_{2,8}^{g} \text{ln}\left[\text{cosh}\left(w_{1,3}^{g} \,J_2^{\bm{A}}\right)\right]                      + w_{2,9}^{g}  \left[\text{cosh}\left(w_{1,6}^{g} \,J_2^{\bm{A}}\right)-1\right] \nonumber \\
        &+ w_{2,10}^{g} \, \text{abs}\left((J_2^{\bm{A}})^2\right) + w_{2,11}^{g} \text{ln}\left[\text{cosh}\left(w_{1,5}^{g} \,(J_2^{\bm{A}})^2\right)\right]                      + w_{2,12}^{g}  \left[\text{cosh}\left(w_{1,8}^{g} \,(J_2^{\bm{A}})^2\right)-1\right] .  \label{Eq-g-initial}
    \end{align}
    Here, $\bm{A}$ should be replaced by either $\GAM$ for the potential associated with the yield surface, $g_{\Phi}$ (Equation~(\ref{Eq-Yield})), and the evolution of plastic strains, $g_1$ (Equation~(\ref{Eq-Dp})), or by $\bar{\bm{\Theta}}$ for the potential $g_2$ (Equation~(\ref{Eq-Dpi})).\\
\FloatBarrier

\section{Automated model discovery}~\label{sec:5}

In the following, numerical results are demonstrated to validate the presented framework of neural networks to discover elasto-plastic behavior. The framework is implemented into the library \textit{TensorFlow}~\citep{abadi2016tensorflow}.
We compute the mean squared error using the first entries of the experimentally observed and predicted Cauchy stress, i.e. 
\begin{align}
    \mathcal{L}(\bm{\sigma}_{11}, \bm{w}) = \dfrac{1}{n_{\text{exp}}} \sum_{n=1}^{n_{\text{exp}}} \dfrac{1}{n_{\text{data}}} \sum_{i=1}^{n_{\text{data}}} {\left( \sigma_{11_i} - \hat{\sigma}_{11_i} \right) }^2 + L_2 \sum_{j=1}^{n_w} w_j^2 + + L_1 \sum_{j=1}^{n_w} |w_j| \label{Eq-loss}
\end{align}
where $\sigma_{11}$ and $\hat{\sigma}_{11}$ refer to the predicted and experimentally obtained stresses, respectively. The number of experiments used for training are denoted by $n_{\text{exp}}$, while $n_{\text{data}}$ are the number of data points per experiment, $n_w$ the number of trainable weights regularized, and $L_1$ and $L_2$ are the regularization factors. Throughout all trainings, the ADAM optimization has been used. If not mentioned otherwise, L2-regularization has been applied with $L_2=0.001$ for all FFNs of the Helmholtz free energies and $L_2=0.0001$ for all potentials, while $L_1=0$. In addition, gradient-clipping has been applied to prevent training against nonphysical values by
$G_{\text{clip}} = G \cdot \text{min} \left(1,\,\sfrac{c}{|G|} \right)$.
Here, $G$ is the initially computed gradient and $c=0.01$ the clipnorm threshold.

\subsection{Specification of network architecture}
The presented framework describes the general design of a generic formulation to discover elasto-plastic material behavior incorporating linear and nonlinear kinematic hardening. As illustrated in Figure~\ref{Fig:RNN}, the constitutive material behavior can be represented using up to six feed-forward networks.\\
To promote sparsity within the networks, $L_1$-regularization is commonly employed to reduce the number of learned weights. While our model is already compact compared to traditional neural network approaches, excessive regularization has, as demonstrated in Section~\ref{sec:5-liao}, led to an overly simplified model. 
To maintain both model performance and interpretability, we refined the network architectures as presented above, with a detailed description provided in the following section. For the elastic Helmholtz free energy, we adopt the form presented in Equation~(\ref{Eq-Psi}).
For the elastic Helmholtz free energy, we choose the form given in Equation~(\ref{Eq-Psi}).
We restrict this work to incompressible material behavior. Therefore, we add a Lagrangian term to the energy that reads $p(I_3^{\ce}I_3^{\bm{C}_p}-1)$. Here, $p$ can be interpreted as the hydrostatic pressure that is computed from the boundary conditions. Considering coaxial loads only, we can define $p$ such that $S_{33}=0$ holds. The energies for linear and nonlinear kinematic hardening are reduced by the squared terms of the first layer, such that we yield 
\begin{align}
    \psi(\bm{C}_p) &= \, w_{2,1}^{\psi_p} \,(\tilde{I}_1^{\bm{C}_p}-3)+ w_{2,2}^{\psi_p} \left( \text{exp} (w_{1,1}^{\psi_p}\,[\tilde{I}_1^{\bm{C}_p}-3])-1\right)  \nonumber \\ \nonumber  
    &+ w_{2,3}^{\psi_p}\, \left((\tilde{I}_2^{\bm{C}_p})^{3/2}-3\sqrt{3}\right)        + w_{2,4}^{\psi_p}  \left(\text{exp} (w_{1,2}^{\psi_p}\,\left[(\tilde{I}_2^{\bm{C}_p})^{3/2}-3\sqrt{3}\right])-1\right)  \\   
    &+ w_{3,2}^{\psi_p}  \left[ {\left(I_3^{\bm{C}_p}\right)}^{ w_{3,1}^{\psi_p}} -1 -  w_{3,1}^{\psi_p} \text{ln} \left(I_3^{\bm{C}_p}\right) \right]  \label{Eq-finalPsi-p}\\
    \psi(\bpe) &= \, w_{2,1}^{\psi_{p_e}} \,(\tilde{I}_1^{\bpe}-3)+ w_{2,2}^{\psi_{p_e}} \left( \text{exp} (w_{1,1}^{\psi_{p_e}}\,[\tilde{I}_1^{\bpe}-3])-1\right)  \nonumber \\ \nonumber  
    &+ w_{2,3}^{\psi_{p_e}}\, \left((\tilde{I}_2^{\bpe})^{3/2}-3\sqrt{3}\right)        + w_{2,4}^{\psi_{p_e}}  \left(\text{exp} (w_{1,2}^{\psi_{p_e}}\,\left[(\tilde{I}_2^{\bpe})^{3/2}-3\sqrt{3}\right])-1\right)  \\   
    &+ w_{3,2}^{\psi_{p_e}} \left[ {\left(I_3^{\bpe}\right)}^{ w_{3,1}^{\psi_{p_e}}} -1 -  w_{3,1}^{\psi_{p_e}} \text{ln} \left(I_3^{\bpe}\right) \right]. \label{Eq-finalPsi-pi}
\end{align}
In addition, we choose an associative evolution law for the evolution of $\bm{C}_p$, such that the potential of the yield criterion and of the plastic stretches are the same, i.e. $g_\Phi = g_1$. Thus, the two potentials $g_1$ and $g_2$ remain to be defined. We describe both by the following reduced formulations, that are
\begin{align}
    g_1(\GAM) &=\, w_{2,1}^{g_1} \,\text{abs}\left(I_1^{\GAM}\right)        + w_{2,2}^{g_1} \text{ln}\left[\text{cosh}\left(w_{1,1}^{g_1} \,I_1^{\GAM}\right)\right] \nonumber \\
              &+ w_{2,2}^{g_1} \, \text{abs}\left((I_1^{\GAM})^2\right)     + w_{2,4}^{g_1} \text{ln}\left[\text{cosh}\left(w_{1,2}^{g_1} \,(I_1^{\GAM})^2\right)\right]    \nonumber  \\
              &+ \tilde{w}_{2,5}^{g_1} \, \text{abs}\left(\tilde{J}_2^{\GAM}\right)         + w_{2,6}^{g_1} \text{ln}\left[\text{cosh}\left(\tilde{w}_{1,3}^{g_1} \,\tilde{J}_2^{\GAM}\right)\right]  \label{Eq-finalYield}\\
    g_2(\bar{\bm{\Theta}}) &=\, w_{2,1}^{g_2} \,\text{abs}\left(I_1^{\bar{\bm{\Theta}}}\right)        + w_{2,2}^{g_2} \text{ln}\left[\text{cosh}\left(w_{1,1}^{g_2} \,I_1^{\bar{\bm{\Theta}}}\right)\right] \nonumber \\
            &+ w_{2,2}^{g_2} \, \text{abs}\left((I_1^{\bar{\bm{\Theta}}})^2\right)     + w_{2,4}^{g_2} \text{ln}\left[\text{cosh}\left(w_{1,2}^{g_2} \,(I_1^{\bar{\bm{\Theta}}})^2\right)\right]    \nonumber  \\
            &+ \tilde{w}_{2,5}^{g_2} \, \text{abs}\left(\tilde{J}_2^{\bar{\bm{\Theta}}}\right)         + w_{2,6}^{g_2} \text{ln}\left[\text{cosh}\left(\tilde{w}_{1,3}^{g_2} \,\tilde{J}_2^{\bar{\bm{\Theta}}}\right)\right] 
\end{align}
Notably, we have scaled the second deviatoric invariant by $\tilde{J}_2 = 3\, J_2$. Thus, we obtain for the corresponding weights the relation $ w_{2,5}^{g} = 3\,\tilde{w}_{2,5}^{g}$ and $w_{1,3}^{g} = 3\,\tilde{w}_{1,3}^{g}$, which is in line with for example~\cite{tacc2023data}.

\subsection{Preprocessing}
All stresses were normalized before being used for training to avoid excessive sensitivity to randomly initialized weights.
In addition, we performed a pretraining of the initial weights. For this, we set all weights to zero and non-trainable and began with training the weights of the elastic Helmholtz free energy. We then sequentially pretrained the weights of the yield function, the linear hardening energy, and the nonlinear hardening energy and the corresponding plastic potential. This strategy ensured that the initial weights were already close to their final values, leading to less error-prone and more stale training processes.\\ 
In particular, a key aspect of this approach was to enforce plastic behavior within the training domain to ensure, that the loss was dependent on the weights of the plastic potential. 
Without this enforcement, a purely elastic iteration step would cause these weights to be automatically set to zero in the next iteration.
Thus, any plastic dependency would be eliminated causing the model to predict purely elastic material behavior.

\subsection{Discovering artificially generated data}
In the following, we investigate the prediction of artificially generated data. For this, we first use data of a model that is included in our network architecture. Subsequently, we aim to find a model that can not be captured by our architecture as it includes tension-compression asymmetry.

\subsubsection{Artificially generated data}
\begin{figure}[h!tbp]
    \includegraphics[scale=0.98]{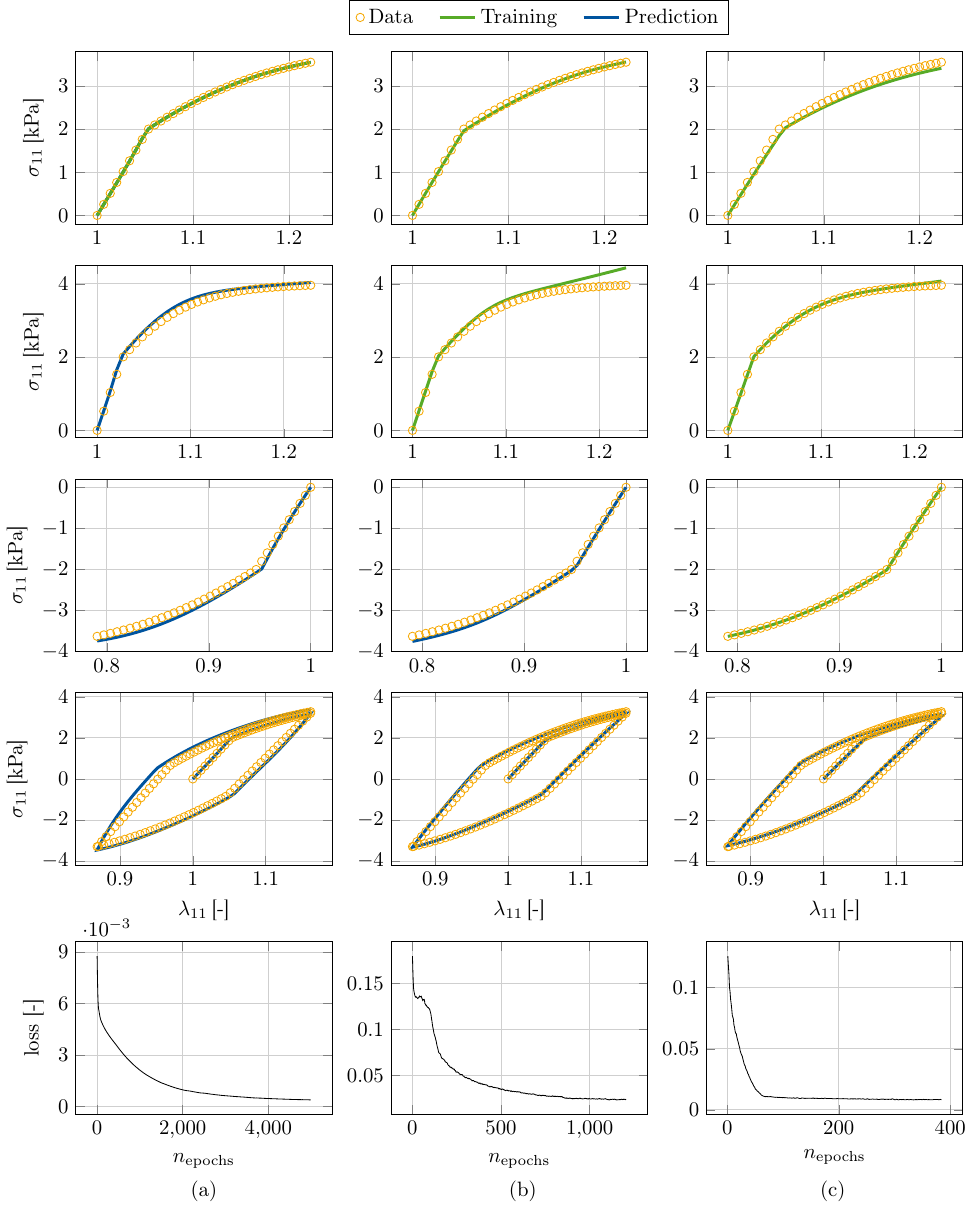}
        \caption{Artificially generated data: Stress strain curves of uniaxial tension (1st row), equibiaxial tension (2nd row), uniaxial compression (3rd row), cyclic loading (4th row), and corresponding loss curves (5th row).
        Trained on (a) uniaxial tension, (b) uniaxial tension and equibiaxial tension, (c) uniaxial tension, equibiaxial tension, uniaxial compression.}\label{Fig-artdata-StressStrain1}
\end{figure}

\begin{figure}[h!tbp]
    \includegraphics[scale=0.98]{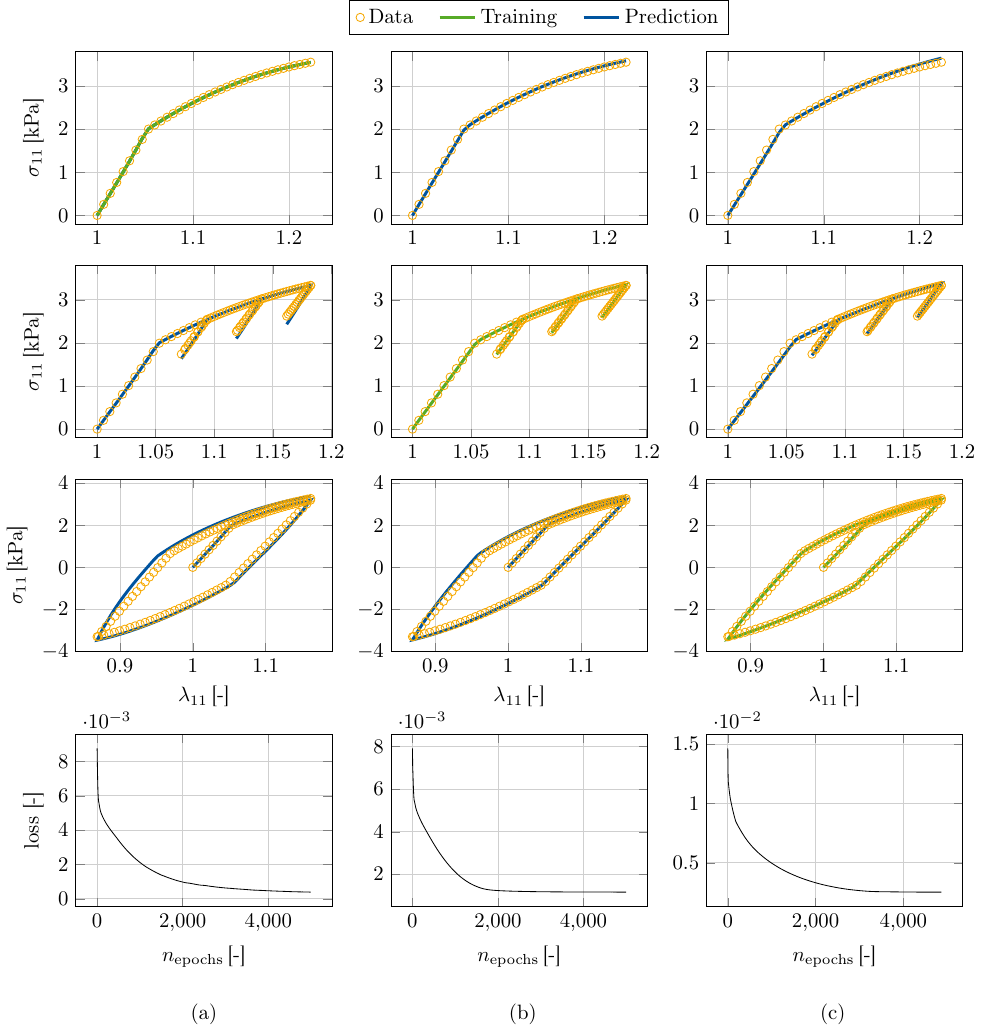}
        \caption{Artificially generated data: Stress strain curves of uniaxial tension (1st row), uniaxial tension with unloading (2nd row), cyclic loading (3rd row), and corresponding loss curves (5th row).
        Trained on (a) uniaxial tension, (b) uniaxial tension with unloading, (c) cyclic loading.}\label{Fig-artdata-StressStrain-2}
\end{figure}

We investigated the performance firstly by training on artificially generated data of an elasto-plastic material with nonlinear kinematic hardening. The underlying data was generated using a Neo-Hookean model type, where the Helmholtz free energy was defined by
\begin{align}
    \psi(\bar{\bm{C}}_e) = \dfrac{\mu}{2} \left(\dfrac{\text{tr}(\bar{\bm{C}}_e)}{\text{det}{(\bar{\bm{C}}_e)}^{1/3}} - 3\right) + \dfrac{K}{4} \left(\text{det}(\bar{\bm{C}}_e) - 1 - \text{ln}[\text{det}(\bar{\bm{C}}_e)]\right) + \dfrac{c}{2} \left( \dfrac{\text{tr}(\bm{C}_p)}{\text{det}(\bm{C}_p)^{1/3}}-3\right)- p \, (I_3^{\bm{C}}-1) 
\end{align}
The last term was added as a Lagrange term where $p$ represented the hydrostatic pressure, derived from the boundary conditions. The yield function was chosen as von Miss type, i.e. given by
\begin{align}
    \Phi(\bar{\bm{\Gamma}}) = \dfrac{3}{2}\, \text{tr}\left(\text{dev}(\bar{\bm{\Gamma}})^2\right) - \sigma_{y_0}^2
\end{align}
with the shear modulus $\mu=12.5\text{ kPa}$, the bulk modulus $K=25\text{ kPa}$ and the yield stress $\sigma_{y_0}=2 \text{ kPa}$. The hardening parameter is $c=8.5 \text{ kPa}$ while the evolution of hardening is described by a plastic potential of $g_2 = \sfrac{b}{2}\,  \text{tr}\left(\text{dev}(\bar{\bm{\Gamma}})^2\right) $ with $b=3 \text{ kPa}$.\\
To evaluate the model's predictive capability, we trained it on different sets of loading conditions. Given that a large amount of data from experiments in reality is often limited due to costs and time, we restricted our data to a realistic set of load sets:
\begin{itemize}
    \item Uniaxial tension~(UT)
    \item Uniaxial compression~(UC)
    \item Equibiaxial tension~(EB)
    \item Uniaxial tension with unloading~(UT-unl)
    \item Uniaxial cyclic loading~(Cyc)
\end{itemize}
These data sets were used in five different combinations to examine the model prediction with different training data. Particularly, the increase in data was investigated by increasing the uniform loads: (UT) $\to$ (UT, EB) $\to$ (UT, EB, UC). Further, we studied the influence of more complex loading paths: (UT-unl) and (Cyc). Each training set was run for a maximum of 5000 epochs, while early stopping was triggered if the loss remained constant for 50 epochs. 
The stress-strain and corresponding loss curves of the increased data of uniform loads are shown in Figure~\ref{Fig-artdata-StressStrain1}, where training on (UT) is shown in (a), training on (UT, EB) in (b), and (UT, EB, UC) in (c). 
Similarly, the results for training on (UT-unl) and (Cyc) are compared in Figure~\ref{Fig-artdata-StressStrain-2}~(b) and~(c), respectively, with training on (UT) in (a) given as a reference. The predicted weights of the yield surface and the Helmholtz free energy accounting for kinematic hardening are summarized in Tables~\ref{Tab:Anal:Weights_Phi} and~\ref{Tab:Anal:Weights_Psi_p_Psi_pi}.\\
\textbf{Results.} \quad
The stress-strain curves indicate a qualitative improvement in the model prediction with increasing training data. While training on uniaxial tension alone provided already satisfactory results for this load case, the load cases used for testing revealed offsets. Expanding the dataset with additional uniform load sets improved the model's prediction as can be seen in Figure~\ref{Fig-artdata-StressStrain1}~(b) and~(c). The best prediction was obtained for complex loading paths, in particular for cyclic loading, see Figure~\ref{Fig-artdata-StressStrain-2}~(b) and~(c).\\
These observations align with the predicted weights. Ideally, the prediction should yield $w_{2,5}^{\Phi} = 0.25$, $w_{2,5}^{g_2} = 1/3$, and $w^{\psi_{p_e}}_{1,1} = 8.5$. From Table~\ref{Tab:Anal:Weights_Phi}, we see that training on UT produces weights close to the expected value of $w_{2,5}^{\Phi}$ but with some noise as there remain some non-zero weights. The minimal number of non-zero weights was observed in the training on cyclic loading. Here, only the weight that is included in the data is non-zero. 
Additionally, analyzing the weights for linear and nonlinear kinematic hardening in Table~\ref{Tab:Anal:Weights_Psi_p_Psi_pi} reveals that the training on cyclic loading led to elimination of all linear hardening terms, ensuring that all hardening effects originated from the nonlinear hardening terms, which was in line with the analytical solution.\\
Interestingly, when comparing the results of the load sets with the greatest variety~--~(UT, EB, UC) and (Cyc)~--~we observed similar predictions but faster convergence during training when multiple load cases where included highlighting the advantage of incorporating diverse load sets in the training data.
Thus, we can conclude from both, qualitatively (stress-strain data) and quantitatively (predicted weights and convergence) that increasing the variety in training data enhances the predictive accuracy of our model. 
Particularly, including loading and unloading paths is crucial for capturing history-dependent material behavior and ensuring realistic hardening predictions.
Cyclic loading data is especially important to distinguish between different hardening effects.
While we did not consider isotropic hardening here, this would even further contribute to the desire of cyclic loading.
%
%
\begin{table}[]
    \begin{tabular}{c | c c c c c}
        \small
        Loadcase   & UT        & UT, EB   & UT, EB, UC & UT-unl    & Cyclic   \\ \hline
    $w_{1,1}^{g_\Phi}$ & 0         & 0        & 0          & 0         & 0        \\
    $w_{1,2}^{g_\Phi}$ & 0         & 0        & 0          & 0         & 0        \\
    $w_{1,3}^{g_\Phi}$ & 0         & 0        & 0          & 0         & 0        \\ \hline
    $w_{2,1}^{g_\Phi}$ & 1.878e-6  & 6.342e-4 & 1.129e-4   & 3.444e-5  & 0        \\
    $w_{2,2}^{g_\Phi}$ & 0         & 0        & 0          & 0         & 0        \\
    $w_{2,3}^{g_\Phi}$ & 7.044e-11 & 5.773e-9 & 1.068e-9   & 4.261e-10 &          \\
    $w_{2,4}^{g_\Phi}$ & 0         & 0        & 0          & 0         & 0        \\
    $w_{2,5}^{g_\Phi}$ & 2.536e-1  & 2.603e-1 & 2.481e-1   & 2.427e-1  & 2.349e-1 \\
    $w_{2,6}^{g_\Phi}$ & 0         & 0        & 0          & 0         & 0       
    \end{tabular}%
    \caption{Artificially generated data: Discovered weights of the yield function (Equation~(\ref{Eq-finalYield})) for the different load cases used for training (UT~--~Uniaxial tension, EB~--~Equibiaxial tension, UC~--~Uniaxial compression, UT-unl~--~Uniaxial tensions with unloading).}\label{Tab:Anal:Weights_Phi}
\end{table}
\begin{table}[]
    \begin{tabular}{c | c c c c c}
        \small
        & UT                   & UT, EB               & UT, EB, UC           & UT-unl               & Cyc                  \\ \hline
        $w_{1,1}^{\psi_p}$   & 0                    & 0                    & 0                    & 0                    & 0                    \\
        $w_{1,2}^{\psi_p}$   & 0                    & 0                    & 0                    & 0                    & 0                    \\
        $w_{2,1}^{\psi_p}$   & 1.510e-2             & 2.853e-1             & 6.461e-2             & 3.933e-2             & 0                    \\
        $w_{2,2}^{\psi_p}$   & 0                    & 0                    & 0                    & 0                    & 0                    \\
        $w_{2,3}^{\psi_p}$   & 4.081e-2             & 1.234e-1             & 5.536e-2             & 9.599e-2             & 0                    \\
        $w_{2,4}^{\psi_p}$   & 0                    & 0                    & 0                    & 0                    & 0                     \\ \hline
        $w_{1,1}^{\psi_{p_e}}$   & 6.063e+0             & 4.644e+0             & 4.648e+0             & 4.848e+0             & 5.197e+0             \\
        $w_{1,2}^{\psi_{p_e}}$   & 4.959e+0             & 3.311e+0             & 3.392e+0             & 3.747e+0             & 4.835e+0             \\
        $w_{2,1}^{\psi_{p_e}}$   & 1.856e-2             & 1.111e-1             & 5.815e-2             & 2.694e-2             & 1.598e-2             \\
        $w_{2,2}^{\psi_{p_e}}$   & 1.121e-1             & 2.869e-1             & 1.863e-1             & 1.524e-1             & 9.283e-2             \\
        $w_{2,3}^{\psi_{p_e}}$   & 4.922e-2             & 1.434e-1             & 1.479e-1             & 6.838e-2             & 5.210e-2             \\
        $w_{2,4}^{\psi_{p_e}}$   & 2.321e-1             & 2.215e-1             & 2.729e-1             & 3.018e-1             & 2.536e-1             \\
    \end{tabular}%
    \caption{Artificially generated data: Predicted weights of the Helmholtz energy accounting for linear kinematic hardening ($\psi_p$) and nonlinear kinematic hardening~($\psi_{p_e}$).}\label{Tab:Anal:Weights_Psi_p_Psi_pi}
\end{table}

\subsubsection{Artificially generated data with tension-compression asymmetry}
\begin{figure}[htbp]
    \includegraphics{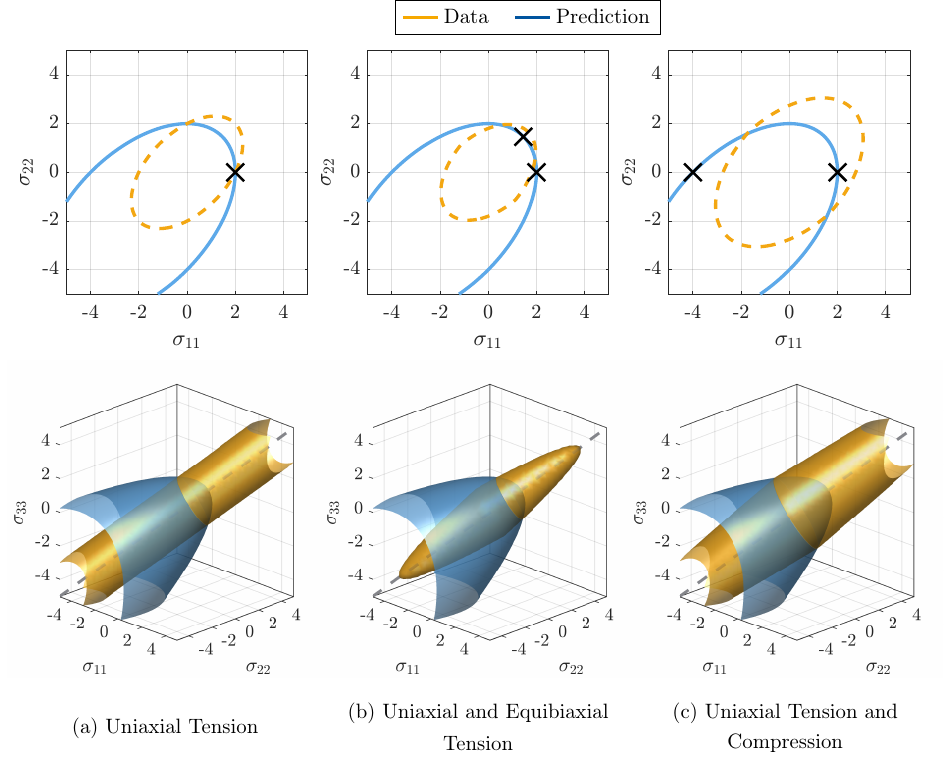}
    \caption{Artificially generated data with tension-compression asymmetry: Underlying yield surface (blue) and predicted yield surfaces (orange) in 2D (upper row) and 3D (lower row) for training on (a) uniaxial tension, (b) uni- and equibiaxial tension, and (c) uniaxial tension and compression (hydrostatic axis is given for reference in gray).}\label{Fig:anal-TCA:Yieldsurface}
\end{figure}

In the previous section, we analyzed the model performance for data that can be captured by our network architecture. To consider cases that could not be captured by the iCANN, we investigated an elasto-plastic material with perfect plasticity and a yield criterion that captured tension-compression asymmetry. This model was based on a compressible Neo-Hookean type energy with a paraboloid yield function following~\cite{tschoegl1971failure}, i.e.
\begin{align}
    \Phi = 3\, J_2 + (\sigma_C - \sigma_T) I_1 - \sigma_c\sigma_T. \label{Eq:Phi-TCA}
\end{align}
Here, the initial yield stress was set to $\sigma_T =2$ MPa in tension and to $\sigma_C=4$ MPa in compression. Tension-compression asymmetry was captured through the sign of the first invariant $I_1$ in Equation~(\ref{Eq:Phi-TCA}). Notably, the activation functions used in our model are non-negative and convex, thereby excluding any dependency of the sign of the inputs and preventing the network from learning the tension-compression asymmetry. 
Again, different combinations of load cases were investigated: (UT), (UT, UC), and (UT, EB). As the main interest lied in investigating the predicted yield criterion, we focused solely on the yield surface prediction by fixing the elastic energy to ensure correct elastic behavior and setting all other weights to zero. 
The discovered yield surfaces are visualized in Figure~\ref{Fig:anal-TCA:Yieldsurface} in 2D in the $\sigma_{11}$-$\sigma_{22}$-plane in the upper line and in 3D in the lower line. The different load cases used for training are shown in (a) for uniaxial tension, (b) uniaxial and equibiaxial tension, and (c) uniaxial tension and compression. The initial yield stresses corresponding to each load set are represented by black markers, providing a visual reference of the training data.
The yield surface of Equation~(\ref{Eq:Phi-TCA}) is visualized in blue and the predicted one in orange.\\
\textbf{Results.} \quad
It can be observed, that for the training using the UT data only, the predicted yield stress matches the reference value accurately. However, the overall shape of the yield surface is not in line with the underlying model (see Figure~\ref{Fig:anal-TCA:Yieldsurface} (a)), where the prediction resembles the von Mises type yield surface. Investigation of the predicted weights given in Table~\ref{Tab:Anal-TCA:Weights_Phi} reveals that the invariant $J_2$ most dominantly affected the result. When the training data set has been increased to include UT and EB, the form of the yield function had changed due to an increased influence of the first invariant and aligned more closely with the training data in the tensile octant (c.f. Figure~\ref{Fig:anal-TCA:Yieldsurface} (b)). Despite this change in the form, the model had failed to capture the asymmetry of the Tschoegel yield function, as the discovered and reference solution still differed significantly in the compressive regime.  
Training on UT and UC resulted in a compromise between tension and compression but still failed to represent the full asymmetry reflected in the yield stresses.
These findings confirm that without explicit mechanisms to capture sign-dependent behavior, the model struggles to learn tension-compression asymmetry. This highlights the need for alternative architectures or additional constraints to improve yield surface predictions for asymmetric materials.
\begin{table}[]
    \begin{tabular}{c | c c c}
        \small
        Loadcase   & UT        & UT, EB   & UT, UC    \\ \hline
        $w_{1,1}^{g_\Phi}$ & 0         & 0         & 0        \\
        $w_{1,2}^{g_\Phi}$ & 0         & 0         & 0        \\
        $w_{1,3}^{g_\Phi}$ & 1.286e-1  & 0         & 0        \\ \hline
        $w_{2,1}^{g_\Phi}$ & 4.024e-7  & 8.562e-2  & 3.709e-2 \\
        $w_{2,2}^{g_\Phi}$ & 0         & 0         & 0        \\
        $w_{2,3}^{g_\Phi}$ & 2.553e-12 & 9.457e-10 & 6.491e-9 \\
        $w_{2,4}^{g_\Phi}$ & 0         & 0         & 0        \\
        $w_{2,5}^{g_\Phi}$ & 2.485e-1  & 2.606e-1  & 1.187e-1 \\
        $w_{2,6}^{g_\Phi}$ & 4.614e-2  & 0         & 4.303e-3

    \end{tabular}%
    \caption{Artificially generated data with tension-compression asymmetry: Discovered weights of the yield function (Equation~(\ref{Eq-finalYield})) for the different load cases used for training (UT~--~Uniaxial tension, EB~--~Equibiaxial tension, UC~--~Uniaxial compression).}\label{Tab:Anal-TCA:Weights_Phi}
\end{table}

\subsection{Comparison to other neural network approaches}\label{sec:5-liao}
\begin{figure}[h!tbp] {(b)};
    \includegraphics{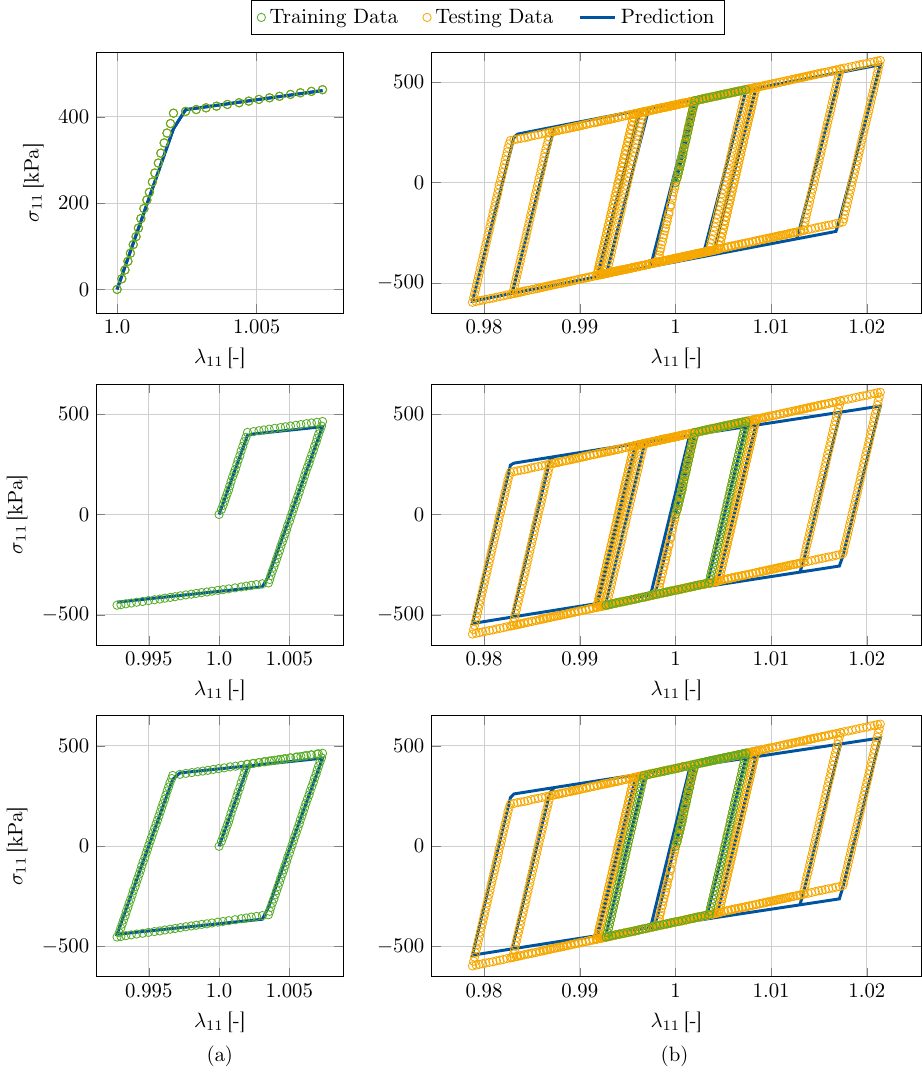}
        \caption{\textbf{Ridge regularization:} Training on data taken from~\cite{liao2025efficient}: Stress-stretch curves for training on the first tensile loading (1st row), tensile and compressive loading (2nd row), and on the first cycle (3rd row). Training data (green marker) and model prediction (blue, solid) is shown on the left and testing data (orange marker) with the model prediction (blue, solid) on the right.}\label{Fig-Liao-StressStrain}
\end{figure}

\begin{figure}[h!tbp]
    \includegraphics{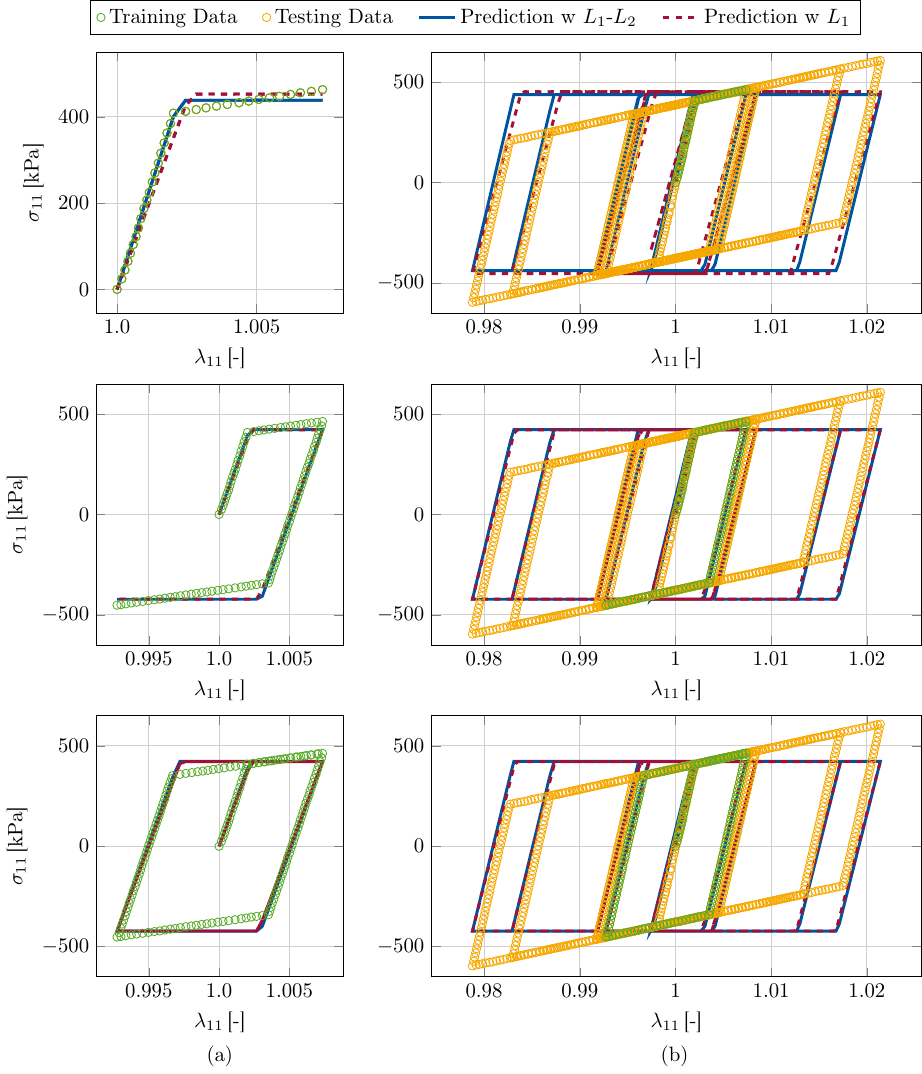}
        \caption{\textbf{Comparison of regularization schemes:} Training on data taken from~\cite{liao2025efficient}: Stress-stretch curves for training on the first tensile loading (1st row), tensile and compressive loading (2nd row), and on the first cycle (3rd row). Training data (green marker) and model prediction for regularization with either $L_2$-regularization for the potentials and $L_1$-regularization for the Helmholtz free energy (blue, solid) or $L_1$ regularization for both (red, dashed) is shown on the left. Testing data (orange marker) is given on the right.}\label{Fig-Liao-StressStrain-l1l2}
\end{figure}

There are numerous other neural network approaches to discover material models including effects of plasticity. One of them is an efficient kernel learning-based constitutive model (EKLC) presented by~\cite{liao2025efficient}. This model is significantly faster than other neural network approaches while effectively capturing elasto-plastic material behavior, including both linear and nonlinear hardening effects.
A key distinction between EKLC and our approach is that our model is formulated within the finite strain theory, making it more flexible to be applied to various materials exhibiting large deformations. Additionally, we do not only learn the yield function and hardening law but also discover the complete constitutive relations, further enhancing the model's adaptability.
As described in Section~\ref{sec:3}, the weights of our feed-forward networks can be interpreted as material parameters due to the structure of the networks.\\
The EKLC model was trained on four cycles and tested on cyclic loading with varying step sizes and showed good agreement with the training and testing data.
Taking the same data set into account, we performed training on a reduced amount of data and investigated our model prediction. Three different training sets were used: tensile loading, tensile and compressive loading, and the first cyclic loading. The corresponding stress-stretch curves are shown in Figure~\ref{Fig-Liao-StressStrain}, where training data is visualized in green, testing data in orange and the model predictions in solid blue lines. The training sets are given in the left plots and testing data are given in the right plots.\\
\textbf{Results.} \quad
It can be observed that the training on the first tension load already yields a good fit to the cyclic data. However, a slight offset in the yield strength was observed at larger stretches, where the transition into the plastic regime was slightly overshot. Overall, the cyclic tension-compression behavior is met quite accurately when training only on a limited amount of data.\\
Contrary to expectations, increasing the training set to include tension-compression and tension-compression-tension loads did not improve the accuracy of the prediction. Instead, the deviation increased for unseen stretch ratios beyond the training range.
Closer examination of the hardening behavior revealed that the offset already existed in the training data predictions and became more pronounced with increasing stretches.
While our model successfully captured the training data, slight offsets in small datasets can lead to huge offsets in increased loading conditions.
 \\ %
\textbf{Impact of regularization techniques.} \quad
Regularization of the weights has proven to be an important tool such as shown by~\cite{mcculloch2024sparse} or~\cite{fuhg2024extreme}. During the training, we explored different regularization methods. While $L_2$-regularization yielded the best results as presented above, other approaches led to too sparse solutions, leading to reduction of inelastic effects in our model. In addition to $L_2$, we investigated $L_1$- or Lasso-regularization, which promotes sparsity of the model parameters by enforcing them to zero.
We applied $L_1$-regularization in the last layers of the Helmholtz free energies to penalize redundant weights. Further, we remained to regularize the last layers of the potentials with either $L_2$-regularization or $L_1$-regularization.
The results, shown in Figure~\ref{Fig-Liao-StressStrain-l1l2}, indicate that while the initial yield stress was predicted accurately, the hardening behavior was not captured. In comparison to Figure~\ref{Fig-Liao-StressStrain}, the regularization led to reduction of the hardening effects and revealed almost perfect plastic behavior. 
Considering the weights of the hardening energy terms given in Table~\ref{Tab:Exp:Weights_Psi_p_Psi_pi}, we can see quantitatively that the hardening behavior was suppressed as the weights were either zero or negligibly small. Only the weights of the first layer, $w_{1,1}$ and $w_{1,2}$, remained active as they were not regularized. However, their influence was minimal as the second-layer weights, responsible for activation, were zero or close to zero.
\begin{table}[]
    \begin{tabular}{c | c c |c c | c c}
        \small
        Trained on & \multicolumn{2}{c}{UT}        & \multicolumn{2}{c}{UT, UC} & \multicolumn{2}{c}{Cyc} \\ \hline
        Regularization     & $L_2$ & $L_1$ & $L_2$ & $L_1$ & $L_2$ & $L_1$ \\ \hline
        $w_{1,1}^{\psi_p}$ & 1.345e+1   & 6.224e+0  & 1.125e+1     & 5.719e+0    & 1.074e+1   & 5.904e+0   \\
        $w_{1,2}^{\psi_p}$ & 1.444e+1   & 9.597e-1  & 1.356e+1     & 9.697e-1    & 1.408e+1   & 9.958e-1   \\
        $w_{2,1}^{\psi_p}$ & 4.376e-3   & 0         & 3.960e-3     & 0           & 3.528e-3   & 0          \\
        $w_{2,2}^{\psi_p}$ & 5.794e-2   & 0         & 4.392e-2     & 1.004e-5    & 3.846e-2   & 1.403e-5   \\
        $w_{2,3}^{\psi_p}$ & 1.123e-2   & 0         & 1.056e-2     & 0           & 9.851e-3   & 0          \\
        $w_{2,4}^{\psi_p}$ & 1.612e-1   & 0         & 1.419e-1     & 2.111e-5    & 1.358e-1   & 0          \\ \hline
        $w_{1,1}^{\psi_{p_e}}$ & 0          & 0         & 0            & 0           & 0          & 0          \\
        $w_{1,2}^{\psi_{p_e}}$ & 0          & 0         & 0            & 0           & 0          & 0          \\
        $w_{2,1}^{\psi_{p_e}}$ & 4.376e-3   & 0         & 3.960e-3     & 1.056e-2    & 3.528e-3   & 0          \\
        $w_{2,2}^{\psi_{p_e}}$ & 0          & 0         & 0            & 0           & 0          & 0          \\
        $w_{2,3}^{\psi_{p_e}}$ & 1.123e-2   & 0         & 1.056e-2     & 0           & 9.851e-3   & 3.234e-6   \\
        $w_{2,4}^{\psi_{p_e}}$ & 0          & 0         & 0            & 0           & 0          & 0         
    \end{tabular}%
    \caption{Comparison of the predicted weights of the Helmholtz energy accounting for linear kinematic hardening ($\psi_p$) and nonlinear kinematic hardening~($\psi_{p_e}$) for regularization with $L_1$-regularization.}\label{Tab:Exp:Weights_Psi_p_Psi_pi}
\end{table}

\subsection{Experimental data}
\begin{figure}[h!tbp]
\includegraphics{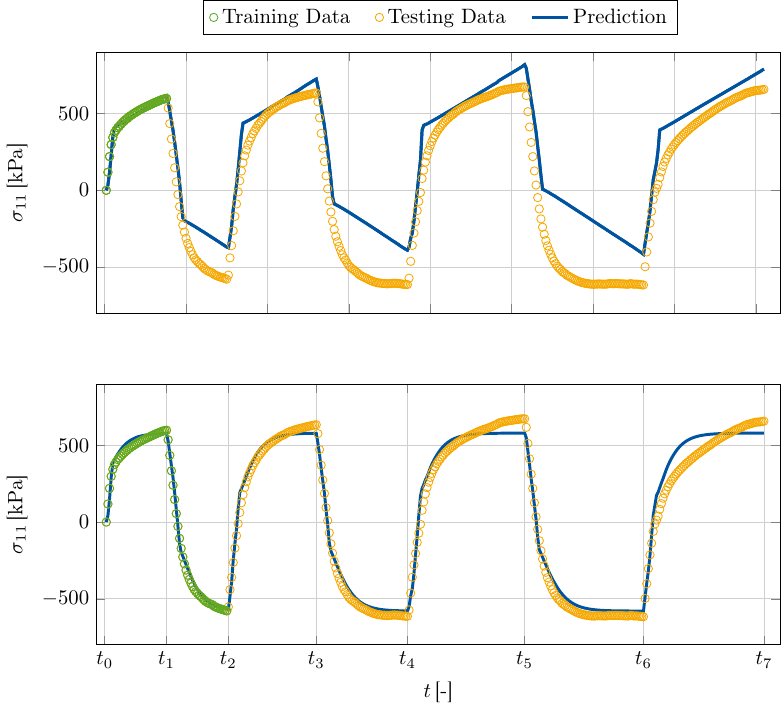}
\caption{Experimental data of DP600 steel taken from~\cite{cao2009experimental}: Stress evolution of time under cyclic tension-compression loading. Training data (green marker), test data (orange marker) and model prediction (blue, solid) for training on the first tension loading path (upper line) and trained on the first tension and compression loading path (lower line).}\label{Fig:exp:Cao}
\end{figure}

To validate our model using experimental data, we examined the cyclic tension-compression tests on DP600 steel conducted by~\cite{cao2009experimental}. Throughout the years, multiple material models have been derived to predict the material response of steels. We aim to use our presented framework to automatically discover the constitutive equations of the underlying experimental data to demonstrate the advantages of our framework.\\
We have conducted two training sets. The first one used only the tensile part of the first cyclic loading path, while the second one was expanded to include both tensile and compressive states from the first cycle. The results are visualized in Figure~\ref{Fig:exp:Cao}, where the upper row shows training on tension data and the lower row training on tension and compression. The data used for training is marked in green, data used for testing by orange markers and the model prediction by solid blue lines.\\
\textbf{Results.} \quad 
The experimental stress curves exhibited nonlinear behavior.
Training on only the tensile part resulted in a good fit for that data, but the predictions deviated significantly for the subsequent loading cycles. The model prediction showed a bilinear behavior, failing to fully capture the nonlinearity observed in the experimental data. The largest deviations occurred in the compressive regime.
Expanding the training data to include both tension and compression led to a smoother transition from elastic to plastic material behavior, improving alignment with the experimental data as can be seen in the lower row in Figure~\ref{Fig:exp:Cao}.
Particularly in the compressive regime, the experimental data was met quite accurately. However, further loading still revealed offsets in predicted versus experimental stress values, particularly in the tensile region, indicating that the model did not fully capture the hardening behavior.

\section{Discussion and limitations}

\textbf{Architecture.}\quad
While our model successfully captured elasto-plastic material behavior with both, linear and nonlinear kinematic hardening effects, several limitations have been observed. 
Due to the current architecture of the potentials, capturing asymmetric material behavior is not possible. Consequently, application to materials exhibiting tension-compression asymmetry remains a challenge. When analyzing the prediction of the Tschoegel yield criterion, it became evident that a variety of loading conditions is necessary in the training data to capture complex yielding behavior. To predict more complex plastic behavior than $J_2$-plasticity, diverse loading must be included. Classical uniaxial tension tests lack sufficient information to define the overall yield surface and evolution of inelastic stretches.\\
Further, our model successfully identified a nonlinear kinematic hardening model for steels. However, the prediction exhibited offsets in increased loading, likely due to isotropic hardening mechanisms not yet included in our framework. Future works should incorporate them to enhance accuracy and a wider set of mechanisms to be captured.

\textbf{Numerical stabilization.}\quad
During training, we observed that training on the Cauchy stress rather than the second Piola-Kirchhoff stress was more stable, in particular for compression data. This effect was particularly pronounced when the initial model predictions were too stiff, as the absolute value of the second Piola-Kirchhoff stress further increases under compression, even for perfect plasticity, whereas the Cauchy stress remains constant. Consequently, we have adopted the Cauchy stress as the primary loss measure.\\
To enhance the training stability, we performed pretraining for the initial weights leading to more efficient training. The initial weights for the pretraining were set randomly. Although one might argue that this procedure introduces a supervised training process, the overall model discovery remains automated, with a preprocessing step in advance. In future works, optimization of this preprocessing should be investigated.
Without the pretraining step, we frequently encountered trainings were the weight optimization led to divergence (NaN values). This was circumvented by applying the pre-processing, normalizing the stresses, and using gradient clipping. However, additional numerical complexity was included due to the Newton-Raphson iteration, requiring evaluation of the gradients of the residual to be minimized to solve for the Karush-Kuhn-Tucker conditions. This additional evaluation might led to problems in the computation of the global gradients. Thus, a derivative-free iteration procedure should be investigated in future work.

\textbf{Effects of regularization.}\quad
Investigation of the effect of regularization methods revealed that $L_1$-regularization led to inaccurate model predictions. Although widely used, particularly as it provides feature selection and leads to sparsity as it enforces weights to be zero, reducing the complexity of the model, it resulted in excessive weight reduction, negatively affecting the prediction of the hardening behavior as observed in the training on data from~\cite{liao2025efficient}. While it can be beneficial for limited data availability, it risks overly simplifying small networks by making them overly sparse.  Its inclusion in the loss function can negatively influence the final prediction, as the reduction in experimental-prediction error might still be outweighed by the penalty introduced by $L_1$-regularization, which increases with the number of weights. Consequently, the hardening effects were essentially eliminated by this regularization approach.
As a result, $L_2$-regularization turned out to be more suitable for the network architecture presented herein. 
To improve the model capacity, future work should consider denser network structures to capture more complex material behavior while reducing the dependency on limited set of neurons used in our current approach. Nevertheless, this study establishes a foundational framework for incorporating plasticity into constitutive artificial neural networks, providing a basis for further advancements and extensions.\\
Using the $L_2$-regularization, we observed that using a smaller regularization factor for the potentials compared to the energies improved accuracy. This choice ensured that the weights of the energies would be penalized more severely, such that they were more likely to be changed to fit the experimental data instead of reducing the weights of the potentials. In contrast, larger regularization factors of the potentials would lead to reduced weights of the potentials. Given our network architecture, where the yield function is divided by the initial yield stress (see Equation~\ref{Eq-Yield}), this could shift the onset of plasticity outside the training data resulting in an entirely elastic model prediction that attempts to approximate the nonlinear stress-strain curve.

\section{Conclusion}
In conclusion, we presented a thermodynamically consistent framework for elasto-plasticity in constitutive artificial neural networks. Dealing with finite strains, mechanisms of linear and nonlinear kinematic hardening were included using a multiplicative decomposition of both the total and the plastic deformation gradient. The constitutive equations of the elastic, linear and nonlinear hardening behavior, as well as the yield surface, and the evolution equations of the inelastic stretches are described by feed-forward networks to be automatically discovered obtaining physically reasonable discovered models due to the design of the networks.\\
To investigate the model's performance, discovery of artificially generated data with a symmetric as well as an asymmetric yield function, experimental data of steels and the comparison to results of other network approaches has successfully demonstrated the capabilities of our model. Overall, most material behavior was predicted quite accurately while the results on the asymmetric yield surface and the experimental data have revealed that the architecture of the potentials, its sparsity and the lack of isotropic hardening leads to discrepancies between the model's prediction and the experimental data. Thus, we should consider the model's architecture and include isotropic hardening. A larger network structure then might require different regularization techniques that should be treated in a new investigation.\\
While numerical instabilities had been circumvented by a pretraining of initial weights, normalization of the stresses prior to the training, gradient clipping, and regularization, future works should investigate the implementation of gradient-free iteration processes to solve the local iteration scheme. However, this remained to be out of the scope of our present work, dealing with the basis of including plasticity into the general design of iCANNs.


\appendix
\section{Declarations}

\subsection{Acknowledgements}   
The authors gratefully acknowledge the financial support of the project SI 1959/12-1 (project number: 466117814) and TRR 280 (project number: 417002380) by the German Research Foundation (Deutsche Forschungsgemeinschaft, DFG). \\
Furthermore, the authors would like to express their gratitude to the help and experience of Jannick Kehls concerning numerical aspects.

\subsection{CRediT authorship contribution statement}
\textbf{Birte Boes}: Conceptualization, Methodology, Software, Formal analysis, Data Curation, Investigation, Validation, Writing~--~Original Draft, Writing~--~Review \& Editing.
\textbf{Jaan-Willem Simon}: Funding acquisition, Writing~--~Review \& Editing.
\textbf{Hagen Holthusen}: Conceptualization, Methodology, Software, Formal analysis, Writing~--~Review \& Editing.

\subsection{Data availability}
Our source code and examples are available under https://doi.org/10.5281/zenodo.1270664 latest after publication.

\subsection{Statement of AI-assisted tools usage}
This document was prepared with the assistance of OpenAI's ChatGPT, which was used for improvement of readibility. The authors reviewed, edited, and take full responsibility for the content and conclusions of this work.


 \section*{References}
 \bibliographystyle{elsarticle-harv} 
 \bibliography{literature.bib}





\end{document}